\documentclass[journal,onecolumn]{IEEEtran}
\IEEEoverridecommandlockouts
\usepackage{fix-cm}
\usepackage{cite}
\usepackage{amsmath,amssymb,amsfonts}
\usepackage{graphicx,color}
\usepackage{comment}
\usepackage{xcolor}
\usepackage{multirow}
\usepackage{url}
\usepackage{textcomp}
\usepackage{adjustbox}
\usepackage{algorithm} 
\usepackage{algpseudocode}   
\usepackage{float}  
\usepackage[hidelinks]{hyperref}

\begin{document}

\title{CO-DEFEND: Continuous Decentralized Federated Learning for Secure DoH-Based Threat Detection\\
}

\author{ 
    \IEEEauthorblockN{Diego Cajaraville-Aboy\textsuperscript{1,*}, Marta Moure-Garrido\textsuperscript{2}, Carlos Beis-Penedo\textsuperscript{1}, Carlos Garcia-Rubio\textsuperscript{2}, Rebeca P. Díaz-Redondo\textsuperscript{1}, Celeste Campo\textsuperscript{2}, Ana Fernández-Vilas\textsuperscript{1}, and Manuel Fernández-Veiga\textsuperscript{1} }\\ 
    \IEEEauthorblockA{ \textsuperscript{1}atlanTTic Research Center -- ICLAB -- Universidade de Vigo, Vigo, 36310, Spain\\ 
    \{dcajaraville,cbeis,rebeca,avilas,mveiga\}@det.uvigo.es }\\ 
    \IEEEauthorblockA{ \textsuperscript{2}Department of Telematic Engineering, Universidad Carlos III de Madrid, Spain\\ 
    \{mamoureg,cgr,celeste\}@it.uc3m.es}\\ \textsuperscript{*}{Corresponding author: dcajaraville@det.uvigo.es}% 
    \thanks{This version of the article has been accepted for publication in Computer Networks after peer review, but is not the Version of Record and may not reflect final copy-editing, formatting, pagination, or corrections. The Version of Record is available online at: \url{https://doi.org/10.1016/j.comnet.2025.111961}. © 2025 The Authors. This manuscript version is made available under the CC BY-NC-ND 4.0 license.} 
}

\maketitle

\begin{abstract}
The use of DNS over HTTPS (DoH) tunneling by an attacker to hide malicious activity within encrypted DNS traffic poses a serious threat to network security, as it allows malicious actors to bypass traditional monitoring and intrusion detection systems while evading detection by conventional traffic analysis techniques. Machine Learning (ML) techniques can be used to detect DoH tunnels; however, their effectiveness relies on large datasets containing both benign and malicious traffic. Sharing such datasets across entities is challenging due to privacy concerns. In this work, we propose CO-DEFEND (Continuous Decentralized Federated Learning for Secure DoH-Based Threat Detection), a Decentralized Federated Learning~(DFL) framework that enables multiple entities to collaboratively train a classification machine learning model for DoH threat detection while preserving data privacy, enhancing scalability and resilience against single points of failure. The proposed DFL framework provides a realistic implementation for DoH threat detection, enabling multiple entities to train their local models online with incoming DoH flows in real-time batches as they are processed -- an approach that fits naturally within modern Internet architectures. This framework adapts four classical machine learning algorithms, Support Vector Machines~(SVM), Logistic Regression~(LR), Decision Trees~(DT), and Random Forest~(RF), for federated scenarios and efficient training. In addition, a key methodological feature of CO-DEFEND is the use of DT and RF as model selection rather than aggregation mechanisms, allowing each participant to retain interpretable and locally optimal decision structures while benefiting from collective updates. We compare our proposed method by using the dataset CIRA-CIC-DoHBrw-2020 with existing machine learning approaches, including more computationally complex alternatives such as neural networks, to demonstrate its effectiveness in detecting malicious DoH tunnels while improving scalability and computational efficiency.
\end{abstract}


\section{INTRODUCTION}

The growing adoption of Domain Name System~(DNS) over Hypertext Transfer Protocol Secure~(HTTPS), known as DoH~\cite{hoffman2018dns}, enhances user privacy by encrypting DNS queries, thereby preventing unauthorized monitoring. However, attackers have exploited DoH to establish covert communication channels, making the detection of malicious activity considerably more challenging.

DoH tunneling has become a significant cybersecurity threat, enabling data exfiltration and Command and Control~(C2) communication. Using encrypted DNS traffic, attackers can maintain persistent access to compromised systems while evading detection by conventional traffic analysis techniques~\cite{bumanglag2020impact}. This covert channel method poses a risk to network security, as it allows malicious actors to bypass traditional monitoring and intrusion detection systems (IDS), making identification and mitigation increasingly challenging.

A major challenge in DoH tunnel detection is the need for large-scale and diverse datasets to train accurate Machine Learning~(ML) models. 
Traditional network security solutions, such as Deep Packet Inspection (DPI)~\cite{wang2021comprehensive}, are less effective against encrypted traffic, limiting their applicability.
To overcome these challenges, intrusion detection systems increasingly rely on ML techniques to classify suspicious patterns in network traffic~\cite{montazeri2020detection}. However, training robust ML models for DoH tunnel detection typically requires centralized data collection, where DNS traffic from multiple sources is aggregated. This centralized approach raises significant privacy concerns, as entities must share sensitive DNS traffic, and it also introduces a single point of failure, making the system more vulnerable to targeted attacks.

Recently, new paradigms have emerged with the purpose of distributing the learning process, thus avoiding the localization of data in a single point and enhancing data privacy. Among them is Federated Learning~(FL)~\cite{McMahan17}, which allows for a collaborative and distributed training process in which each participant trains a local ML model with only its own data. Subsequently, participants share their ML models with a central server that aggregates them to compute a global model. This iterative process also presents a decentralized approach, Decentralized Federated Learning (DFL), where the presence of a central server disappears to favor scalability and robustness against failures in large, sparse, and dynamic scenarios~\cite{Hallaji24}.

Besides that, in certain real-world deployments, it is not common to have a complete and static dataset available to train an ML algorithm from the beginning. In such cases, data are produced continuously, and ML algorithms are trained as new samples arrive, for instance, in the context of sensor data. In addition, the distribution of this data may vary over time due to diurnal patterns, policy changes, or, most important in our case, adversarial behavior. This streaming and non-stationary setting is particularly characteristic in DNS and DoH fields, where data arrives temporally in sequential order and generated from different end users. Consequently, processing this data requires methods such as continuous training from successively arriving batches rather than assuming a single, monolithic training set.

To overcome the current limitations of ML schemes for DoH tunnel detection, we propose CO-DEFEND, a real-time adaptive framework that enables collaborative training for detection of malicious DoH activity through DFL. In realistic deployments, DoH traffic is distributed across different servers, meaning that no single entity can access all flows. Each participant only observes the traffic it receives, making centralized approaches impractical. For this reason, we design CO-DEFEND to operate in such environments by enabling decentralized collaborative learning, where raw data remains local. Furthermore, since DoH traffic patterns are highly dynamic, our framework supports the continual arrival of data through batch updates, ensuring that models can adapt online to evolving threats. CO-DEFEND is thus conceived as a scalable, resilient, lightweight, and easily integrable solution for modern Internet infrastructures. Our contributions can be summarized as follows:
\begin{itemize}
    \item We introduce CO-DEFEND, a novel DFL-based framework designed for the collaborative training of DoH malicious traffic detection models across multiple entities under continual data arrival, emphasizing scalability, system robustness and data privacy in realistic deployments where each entity only observes its own traffic.
    \item We adapt classical ML algorithms, like Support Vector Machines~(SVM), Logistic Regression~(LR), Decision Trees~(DT), and Random Forest~(RF), for the lightweight detection of malicious DoH traffic in continual federated scenarios, and efficient model updates compared to heavy neural network alternatives. Moreover, we design a selection-based strategy for DT and RF, instead of aggregation-based methods, highlighting a practical and interpretable alternative for DFL with non-parametric learners.
    \item We compare different deployment settings (non-federated, centralized and decentralized federated learning) to assess their effectiveness in DoH tunnel detection, focusing on the trade-offs between detection performance, communication and computational overheads due to the collaborative training, as well as their ability to adapt to dynamic, real-world network scenarios.
    \item We employ and adapt a state-of-the-art simulation tool to validate our approach, along with libraries supporting continual batch training with the selected ML algorithms on real-time DoH traffic. In addition, we provide open-source simulator extensions and scripts to ensure full reproducibility of our experiments.
    \item We compare our proposed method with existing centralized FL approaches in the literature, demonstrating that it achieves competitive results in detecting malicious DoH tunnels while providing benefits of enhanced data privacy, scalability and cost reduction.
\end{itemize}

The remainder of the paper is organized as follows: Section~\ref{sec:background} presents an overview of the fundamental concepts and methodologies relevant to this study. Section~\ref{sec:related_work} provides a review of the state of the art related to DoH tunnel detection, FL frameworks, and continuous learning in federated environments. The proposed DFL scenario, methodology, and the definition/adaption of the collaborative training of the chosen ML algorithms are detailed in Section~\ref{sec:methodology}. Section~\ref{sec:implementation} presents the experimental setup of the validation scenarios, and Section~\ref{sec:evaluation} contains the evaluation and discussion of the obtained results. Finally, Section~\ref{sec:conclusion} summarizes the conclusions drawn from our findings and outlines some directions for future research.

\section{BACKGROUND}
\label{sec:background}

This section provides a concise overview of the key concepts and methodologies underpinning this study, including DNS security challenges, federated learning (and its decentralized variant), and incremental learning. We begin by examining the security challenges associated with the DoH communication protocol, focusing on its potential misuse for covert communication. Next, we introduce FL framework and its extension to DFL, emphasizing their advantages in privacy-preserving network security applications. Additionally, we explore incremental learning, a technique that enables models to continuously adapt to evolving attack patterns.

\subsection{DNS Security Threat}
\begin{figure}
    \centering
    \includegraphics[width=0.8\linewidth]{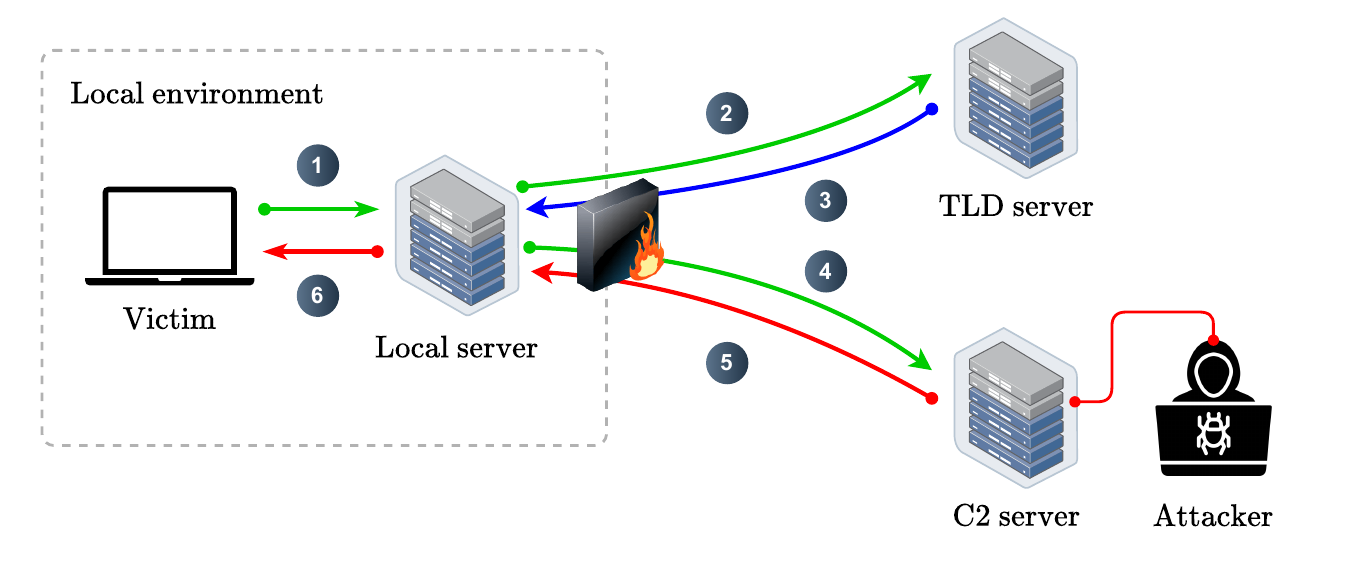}
    \caption{Diagram of the different stages of DNS tunnels functionality}
    \label{fig:diagram_tunnels}
\end{figure}

DNS~\cite{mockapetris1987domain} is a fundamental protocol in pervasive communications, facilitating various functionalities, including service discovery and name resolution. However, the DNS protocol can be exploited as a tunneling mechanism, posing a significant security threat to ubiquitous networks~\cite{hynek2022summary}. 
DNS tunneling exploits the DNS protocol to encapsulate other protocols and transmit data, often bypassing network restrictions for malicious purposes. A DNS tunnel enables bidirectional communication between a client and a server by encapsulating data within DNS packets. 
As shown in Figure~\ref{fig:diagram_tunnels}, the process involves configuring a malicious DNS server, embedding data within DNS queries, and transmitting them to the attacker's server. Once received, the server decodes the data, enabling activities such as data exfiltration, command execution, or network reconnaissance.

A variety of tools exist to support the creation of DoH tunnels, facilitating the transmission of malicious traffic within encrypted DNS connections~\cite{merlo2011comparative}. The primary function of these tools is to establish covert data tunnels by encapsulating traffic within DNS queries and transmitted over HTTPS. Additionally, some tools allow attackers to create DoH tunnels by operating a DoH proxy. Notable examples of such tools include Iodine~\cite{iodine}, dns2tcp~\cite{dns2tcp}, and dnscat2~\cite{dnscat2}.

The covert nature of DNS tunneling poses a significant threat to cyber-physical systems (CPS), as attackers leverage this method to stealthily extract sensitive information or transmit malicious commands. By evading traditional IDS, including behavioral analysis techniques, these attacks compromise the security and integrity of network systems. 
One method for detecting DNS tunnels involves analyzing the content of DNS traffic. However, with the introduction of DoH~\cite{hoffman2018dns}, new challenges have emerged due to the encryption of DNS queries, which limits deep packet inspection. While DoH mitigates certain security vulnerabilities in traditional DNS~\cite{schmid2021thirty}, it also renders conventional detection techniques obsolete, as encryption prevents direct content analysis of DNS traffic~\cite{bumanglag2020impact}.

Discovered in 2019, the Godlua Backdoor~\cite{turing2019analysis} was the first known malware to utilize DoH as a covert communication channel, enabling the concealment of malicious traffic within DNS queries. This development provided concrete evidence of the operational use of DoH, a protocol originally designed to enhance privacy, as a tool for evading conventional network security mechanisms. 
Building on this foundational threat, in 2020, APT34~\cite{apt} became the first documented Advanced Persistent Threat~(APT) group to leverage DoH specifically for data exfiltration. This advancement was critical because it showcased the evolution of DoH-based attacks from simple command and control to the much more insidious act of stealing sensitive information. The use of DoH by an APT group underscored the protocol's effectiveness in bypassing deep packet inspection and other network-level security measures, as the encrypted DoH traffic blends in with legitimate HTTPS traffic.

The importance of these examples lies in their demonstration of the dual nature of DoH, while intended to enhance user privacy and security through encrypted DNS, its adoption also introduced a sophisticated new vector for cyberattacks.

\subsection{(Decentralized) Federated Learning}

FL framework~\cite{McMahan17} has emerged in recent years as an alternative to classical ML, in which training data is aggregated in a central server. Motivated by the emergence of paradigms such as the Internet of Things~(IoT) or computing approaches such as cloud-continuum~\cite{Bittencourt18}, the classical centralized FL~(CFL) consists of collaborative training of an ML algorithm, where each client trains its local model with its own dataset. Then, when the clients finish this training, the local models are sent to a central server,  which is responsible for aggregating them into a single global model (which is sent back to the clients). The federation process is iterative, repeating the above process over a series of rounds starting with the global model from the previous round.  

Although FL has great advantages as a result of learning without sharing data, which increases privacy by not sending data to a processing point and saves bandwidth, it still has several drawbacks~\cite{Wen23,Mothukuri21}. These are frequently discussed in the literature and include both security and privacy issues, like Byzantine attacks by malicious nodes~\cite{Fang20,Shi22} or dealing with delays due to straggler devices~\cite{Schlegel23}, among others. Additionally, even though raw data remain local in FL, model updates can still leak sensitive information (e.g., via gradient inversion~\cite{Geiping20} or membership inference~\cite{Nasr19}). Common countermeasures in the state of the art include cryptographic secure aggregation, differential privacy~\cite{elgabli2025novel}, and hybrid schemes that combine coding with secure aggregation. These defenses typically trade some utility and add communication/computation overhead trade-offs that should be weighed against application requirements. 

Recently, a new paradigm called DFL~\cite{Yuan24} has emerged as a new approach to the centralized version. In DFL, the central server responsible for model aggregation disappears and delegates this procedure to some clients in the learning network. Now, the clients propagate over a peer-to-peer overlay via pairwise or neighborhood exchanges, resulting in different communication topologies. This approach provides a more robust scheme against single-point failures and is more scalable because it does not depend on the central server. DFL also has a number of drawbacks that have received attention in the literature, such as increased model communication overhead~\cite{Ye22}. A fully-connected DFL broadcast is conceptually simple but incurs quadratic communication per round, $\mathcal{O}(n^2)$, which is impractical at scale. Lightweight communication protocols are commonly considered in the literature for dealing with this drawback. Gossip protocol~\cite{Boyd06} restrict each node to exchange with a fixed number of peers per round, reducing total per-round communication to $\mathcal{O}(n)$ while still achieving rapid information mixing on dense overlays. This mechanism can disseminate any model representation. Mixing/convergence speed is governed by the spectral gap of the overlay communication topology; dense, well-connected (cross-silo) graphs mix rapidly, for example.

Security challenges are also present in DFL, even to a greater extent than in CFL. Removing the central aggregator in DFL also removes a potential gatekeeper that can perform server-side sanity checks on client updates. As a result, DFL increases the attack surface for model-poisoning because adversarial peers can disseminate manipulated updates peer-to-peer. To mitigate these threats, researchers proposed different Byzantine-robust aggregation techniques~\cite{cajaraville2024byzantine}.

The concepts previously explained serve as the foundation for our proposed framework, which leverages DFL techniques to enhance the collaborative learning of the detection of malicious DoH tunnels while ensuring scalability, robustness and data privacy.

\section{RELATED WORK}
\label{sec:related_work}

This section provides an overview of existing research on DoH tunnel detection, as well as the application of federated learning in network analysis and streaming/continuous learning methods. 

\subsection{DNS Tunnels}

In recent years, the focus on privacy has intensified as a result of concerns about data security and user privacy. The detection of malicious DoH tunnels attracted considerable attention, driven by the growing threat of encrypted DNS traffic exploited for covert communications. The enhanced confidentiality and integrity of DNS queries alone do not suffice to mitigate all security risks. Research challenges for DoH abuse are presented in~\cite{hynek2022summary}.

The papers in Table~\ref{tab:related_works} focus on DoH tunnel detection using various ML techniques, all of which utilize the ``CIRA-CIC-DoHBrw-2020'' dataset by MontazeriShatoori et al.~\cite{montazeri2020detection}.

\begin{table}[ht]
    \centering
    \caption{Related works of DoH tunnel detection. }
    \label{tab:related_works}
    \renewcommand{\arraystretch}{0.8}
    \resizebox{0.6\linewidth}{!}{%
   \begin{tabular}{|c|c|c|c|c|}
\hline
\textbf{Reference} & \textbf{Year} & \textbf{\# features} & \textbf{ML} & \textbf{Accuracy} \\ \hline
\cite{montazeri2020detection} & 2020 & 28 & RF & 99.9 \% \\
\cite{banadaki2020detecting} & 2020 & 34 & LightGBM, XGBoost & 100\% \\
\cite{singh2020detecting} & 2020 & 31 & RF, GB & 100\% \\
\cite{behnke2021feature} & 2021 & 27 & LightGBM & 100\% \\
\cite{alenezi2021classifying} & 2021 & 31 & XGBoost, RF & 99.22\% \\
\cite{jha2021detection} & 2021 & - & DeepFM & 99.5\% \\
\cite{zebin2022explainable} & 2022 & 29 & RF & 99.98\% \\
\cite{abu2023lightweight} & 2023 & 6 & Adaboost & 100\% \\
\cite{alzighaibi2023detection} & 2023 & - & RF, DT & 99.99\% \\
\cite{mitsuhashi2023malicious} & 2023 & 28 & CatBoost & 99.9\% \\
\cite{niktabe2024detection} & 2024 & 4 & LR & 95.35\% \\ \hline
\end{tabular}
     }
\end{table}

The studies on DoH tunnel detection employ a variety of machine learning algorithms. 
Banadaki~\cite{banadaki2020detecting}, Behnke et al.~\cite{behnke2021feature}, and Alenezi and Ludwig~\cite{alenezi2021classifying} use LightGB and XGBoost, while MontazeriShatoori et al.~\cite{montazeri2020detection}, Singh and Roy~\cite{singh2020detecting}, Zebin et al.~\cite{zebin2022explainable} and Alzighaibi~\cite{alzighaibi2023detection} achieve high accuracy with RF.

Several works also incorporate feature selection techniques~\cite{niktabe2024detection}, improving accuracy by removing irrelevant features~\cite{behnke2021feature} and using correlation-based feature analysis~\cite{jha2021detection}. Additionally, Abu et al.~\cite{abu2023lightweight} leverage Principal Component Analysis (PCA) for feature selection to enhance their Adaboost classifier’s performance.

Niktabe et al.~\cite{niktabe2024unveiling} applied interpretability models to identify patterns in malicious traffic associated with DoH tunnels. In another study, the same authors~\cite{niktabe2024detection} analyzed malicious DoH traffic and proposed an approach that incorporates feature selection, achieving an accuracy of 95.35\% using LR.

Despite the variety of machine learning algorithms and feature selection methods used for this dataset, none of the reviewed studies employ distributed learning approaches for DoH tunnel detection. This reliance on centralized entities poses problems for data privacy and increases communication overhead and delay due to inference requests at distant servers in data centers.

\subsection{Continuous Training in Streaming and Federated Settings}

In real-world deployments, data arrive over time and its distribution can variate (which is called concept drift), so models must be updated as new data is received. One of the first work in consider this type of scenarios was~\cite{domingos2000mining} where authors propose a online Hoeffding-bound decision tree with bounded memory. Building in this technique, \cite{bifet2010moa} introduced MOA framework to implement stream-learning algorithms at scale, including Hoeffding-tree families. Also, some works such as~\cite{gama2014survey} have explored strategies to adapt concept drift phenomenon.

In practice, continuous learning if often handled via updating mini-batches (e.g., sliding windows) in order to reflect data recency meanwhile bounding computation. For example, ADWIN is proposed in~\cite{bifet2007learning} as an adaptive windowing method that detects changes and triggers model refresh based on statistically significant differences between subwindows. By using Hoeffding Trees methods, the work in~\cite{manapragada2018extremely} introduced a Hoeffding-anytime variant to improve responsiveness on evolving streams.

Additionally, the same need arises in federated scenario  when some clients do not own a static dataset at the beginning of the training process. Addressing this, \cite{marfoq2023federated} formulated FL framework for continuous data arrival and proposed a weighted empirical-risk objective so clients can contribute model updates built from recent local data between rounds, including theoretical guidance. For non-stationary considerations, the work in~\cite{jothimurugesan2023federated} proposed client-clustering mechanisms driven by local drift detection to adapt aggregation process when nodes experience different levels of drift asynchronously. For realistic deployments, \cite{wu2024effective} presented a framework for IIoT streaming data that introduced timely round scheduling and lightweight updates, which can sustain accuracy under bandwidth and resource constraints. A complementary research line targets incremental federated learning, aiming to retain prior knowledge while adapting to new tasks/classes. For example, in~\cite{GONZALEZSOTO2024110137}, it is proposed a distributed and collaborative machine learning framework for low-resource IoT devices that provides an incremental prototype-based learning algorithm with random-based model exchange protocols and novel prediction and prototype generation methods.

Together, these works motivate our choice of sequential batch-wise training on continually arriving data, in order to track time-varying DoH patterns without storing full history. This is combined with decentralized learning techniques to keep models aligned with time-varying patterns of DoH streaming: it prioritizes recency, reduces staleness under distribution shifts, and keeps learning tractable without retaining the entire stream.

\subsection{Federated Learning for Traffic Network Analysis}

Several studies have been conducted on the incorporation of FL techniques in various domains, such as health, finance, smart grid, industrial automation, and IIoT~\cite{Qayyum22,Long20,Su21,Zhang21}. Through this approach, it is possible to train ML algorithms collaboratively, guaranteeing the privacy of sensitive information and demonstrating their effectiveness in disease prediction, financial data analysis, or industrial process improvement. However, in comparison, its application to the analysis and detection of malicious traffic in DoH requests has not been widely explored, despite its importance for security in various applications.

Some previous works used FL in the context of network traffic analysis, primarily in network traffic anomaly detection. Doriguzzi et al.~\cite{Doriguzzi24} proposed FLAD (adaptive FL Approach to DDoS attack detection), an approach that optimizes the FL process in cybersecurity by assigning more computing resources to participants with harder-to-learn attack profiles. This enabled constant model refinement while keeping test data private and maximizing both convergence and accuracy for DDoS attack detection. In~\cite{Zhao19}, the authors introduced MT-DNN-FL, which enhanced the FL framework by incorporating a multi-task deep neural network. This approach simultaneously detected network anomalies, identified VPN (Tor) traffic, and classified network traffic, alleviating the data limitation problem while reducing the training burden. 

For the particular case of DoH traffic, the existing approaches are limited. Recently, researchers started exploring FL scenarios for DoH tunnel detection. Li et al.~\cite{li2022detecting} presented a classifier to detect DoH tunnels using FL to train a CNN. However, ensuring the robustness of these models remains challenging, and there is a lack of studies that effectively address these issues. 

Within the research field of FL, the DFL variant received less attention compared to the classical centralized version, despite its advantages in terms of robustness and scalability. The fact that nodes work more autonomously and share knowledge with each other without intermediaries makes this paradigm much more suitable for areas such as Flying Ad-Hoc Networks (FANETs), vehicular scenarios, Industry 4.0, and blockchain, among others~\cite{Beltran23}. For example, Pokhrel et al.~\cite{Pokhrel20} proposed an autonomous blockchain-based FL design for privacy-aware vehicular communication, where on-vehicle machine learning updates were decentralized and verified using blockchain. However, to the best of our knowledge, no studies have explored the use of the DFL paradigm for network traffic analysis.

In this study, we bridge the gap in DFL applications for encrypted traffic analysis by proposing a framework tailored to the unique challenges of DoH tunnel detection. While prior (e.g.,~\cite{Doriguzzi24, Zhao19, li2022detecting}) works demonstrated the potential of FL in network security, their focus has largely been on centralized architectures or non-encrypted traffic scenarios. Our framework combines data streaming learning with decentralized federated updates, enabling adaptive anomaly detection in encrypted network environments. Following sections detail our methodology, experimental validation, and comparative analysis, demonstrating how decentralized federated updates with continuous data arrival overcome these challenges without compromising detection accuracy.

\subsection{Federated Adaptation of ML Models}

The aggregation process in the FL paradigm is a key step, as it enables nodes to learn from others in the learning environment. This process must be tailored to the specific type of ML model being trained, since not all model classes lend themselves equally well to parameter sharing. While parametric models are widely studied in the state of the art (e.g., neural networks), non-parametric models such as decision trees pose more complex challenges. For example, parametric ML algorithms (including SVM or LR) can be represented as fixed-sized parameter vectors. Aggregation depends only in parameter averaging; each node averages its parameters with those received from peers -- a peer-to-peer adaptation of FedAvg~\cite{McMahan17}. On the other hand, adapting tree-based algorithms to FL scenarios is challenging because they are non-parametric and lack aligned parameters for averaging. The literature clusters into:
\begin{itemize}
  \item \emph{Vertical FL for trees (secure split-finding):} Participants hold disjoint features of the same samples; protocols compute split histograms/gradients via secure aggregation, homomorphic cryptography, or multi-party computing. This provides strong privacy guarantees but incurs significant cryptographic overhead.
  \item \emph{Horizontal FL for ensembles (same features, disjoint samples):} Approaches include \textbf{ensemble voting}, where trained trees are exchanged and predictions are aggregated by majority voting (modest communication but brittle under non-IID as weak trees propagate~\cite{Truex19}), and \textbf{selection/pruning}, where nodes exchange trained trees but retain only those validated as effective locally, creating selection pressure toward generalizable rules and showing robustness under non-IID~\cite{Markovic22,Souza20,Wu20}.
  \item \emph{Structural/rule merging:} Attempts to align or merge subtrees and rule sets. However, alignment across heterogeneous trees is combinatorial and often brittle in non-IID settings; computational costs grow with depth and size.
\end{itemize}

Our work intersects three different areas of distributed machine learning: decentralized orchestration, gossip-based communication, and the federation of parametric and non-parametric models. Specifically, CO-DEFEND proposal targets a horizontal, cross-silo, non-IID setting and uses a unified pairwise gossip protocol for communication-efficient exchange. It applies the most suitable aggregation per model class: peer-to-peer model averaging for parametric algorithms, and selection/pruning methods for tree-based algorithms. This avoids the heavy cryptography of vertical FL and the brittleness of naïve voting/structural merging for trees. The combination is lightweight, scalable, and effective across diverse provider distributions. Table~\ref{tab:context_co_defend} provides a comparative analysis of relevant federated learning paradigms discussed in the literature.

\begin{table*}[ht]
\centering
\caption{Comparative analysis of Federated Learning paradigms, positioning \textsc{CO-DEFEND}. Communication cost is per round for \(n\) nodes and model size \(\lvert\theta\rvert\).}
\label{tab:context_co_defend}
\footnotesize
\renewcommand{\arraystretch}{1.2}
\setlength{\tabcolsep}{4pt}
\begin{adjustbox}{max width=\textwidth}
\begin{tabular}{|l|l|l|l|l|p{7.4cm}|}
\hline
\textbf{Paradigm} & \textbf{Orchestration} & \textbf{Aggregation Strategy} & \textbf{Comm./round} & \textbf{Non-IID robustness} & \textbf{Key trade-off / rationale} \\ \hline

\multicolumn{6}{|l|}{\textbf{1. Centralized baseline}} \\ \hline
FedAvg~\cite{McMahan17} & Central server & Parameter averaging & \(\mathcal{O}(n\cdot \lvert\theta\rvert)\) & Moderate & Simplicity and control vs.\ single point of failure and privacy risks. \\ \hline

\multicolumn{6}{|l|}{\textbf{2. Decentralized approaches for parametric models}} \\ \hline
Fully-connected DFL & P2P broadcast & Parameter averaging & \(\mathcal{O}(n^{2}\cdot \lvert\theta\rvert)\) & Moderate & Fully decentralized but communication overhead is prohibitive at scale. \\
Gossip DFL~\cite{Boyd06} & P2P pairwise & Parameter averaging & \(\mathcal{O}(n\cdot \lvert\theta\rvert)\) & High & Scalable and robust, but convergence speed depends on network topology. \\ \hline

\multicolumn{6}{|l|}{\textbf{3. Federated tree strategies (horizontal FL)}} \\ \hline
Ensemble voting~\cite{Truex19} & P2P / Server & Majority voting & \(\mathcal{O}(n\cdot \lvert\theta\rvert)\) & Low / Sensitive & Simple to implement but brittle; weak learners can degrade performance. \\
Selection \& pruning~\cite{Markovic22} & P2P / Server & Performance-based selection & \(\mathcal{O}(n\cdot \lvert\theta\rvert)\) & High & Adaptive and robust, as only high-performing models propagate. \\ \hline

\multicolumn{6}{|l|}{\textbf{4. \textsc{CO-DEFEND} framework (our proposal)}} \\ \hline
\textit{Parametric Models} & P2P gossip & Parameter averaging & \(\mathcal{O}(n\cdot \lvert\theta\rvert)\) & High & Leverages the efficiency of gossip for standard parametric model updates. \\
Federated Trees & P2P gossip & Selection/pruning & \(\mathcal{O}(n\cdot \lvert\theta\rvert)\) & High & Combines gossip efficiency with an aggregation method robust for non-IID tree models. \\ \hline

\end{tabular}
\end{adjustbox}
\end{table*}

\section{METHODOLOGY}
\label{sec:methodology}

This section details the configuration of our federation scenario, including the selection and adaptation of the learning algorithms employed by the participating entities. We also describe the technical aspects of distributed federated and data streaming training that enable effective model learning in a decentralized environment.

\subsection{System Configuration}

\begin{figure}
    \centering
    \includegraphics[width=0.39\linewidth]{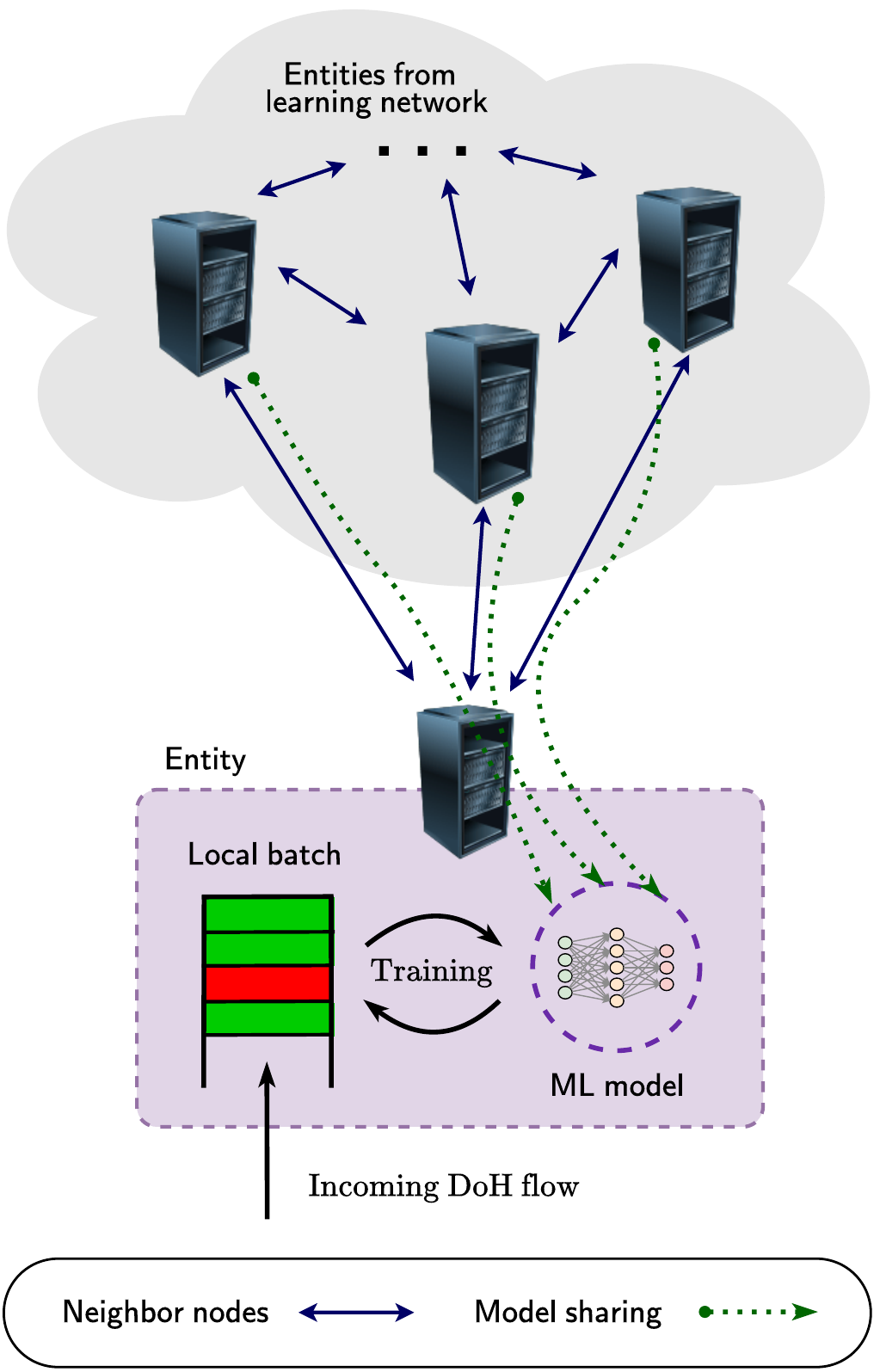}
    \caption{Decentralized Federated Learning framework scenario}
    \label{fig:dfl_scenario}
\end{figure}

We consider a distributed learning scenario involving multiple entities (i.e. a cross-silo scenario), each utilizing different DNS provider and collaborating to detect malicious DoH tunnels. Since DNS queries are encrypted in DoH, deep inspection of the packet is not feasible; instead, intrusion detection relies on traffic pattern analysis. Each entity deploys an IDS with the primary objective of identifying intrusions through network traffic analysis and detecting malicious DoH tunnels.
For this purpose, we employ a DFL approach for collaboratively training a classification ML algorithm. This approach has three main objectives: (i) to strengthen the learning process by using data from different DNS providers; (ii) to avoid data privacy issues by not having to centralize the data location; and (iii) to avoid single point failure issues by not centralizing the federation process. Furthermore, CO-DEFEND is a data streaming framework in which local models are continuously updated as new data arrives at each entity, allowing the system to adapt dynamically to emerging malicious DoH tunnel patterns and gradual shifts in traffic statistics or behavioral features over time (i.e., gradual concept drift).

Figure~\ref{fig:dfl_scenario} illustrates the proposed scenario, depicting the entities participating in the collaborative learning network, each utilizing different DNS providers. The diagram highlights the key components involved in the DFL process, demonstrating how entities contribute to model training.

The topology of the DFL scenario is modeled as a static undirected graph $G = (\mathcal{V}, \mathcal{E})$ where $\mathcal{V}$ represents the nodes participating in the learning process as vertices, which are interconnected by the set of links $\mathcal{E} \subseteq \mathcal{V} \times \mathcal{V}$. Specifically, this is a peer-to-peer (P2P) topology, where each of the entities in the network can communicate with any other entity (motivated by the cross-silo nature of the scenario).

\begin{figure}
    \centering
    \includegraphics[width=0.5\linewidth]{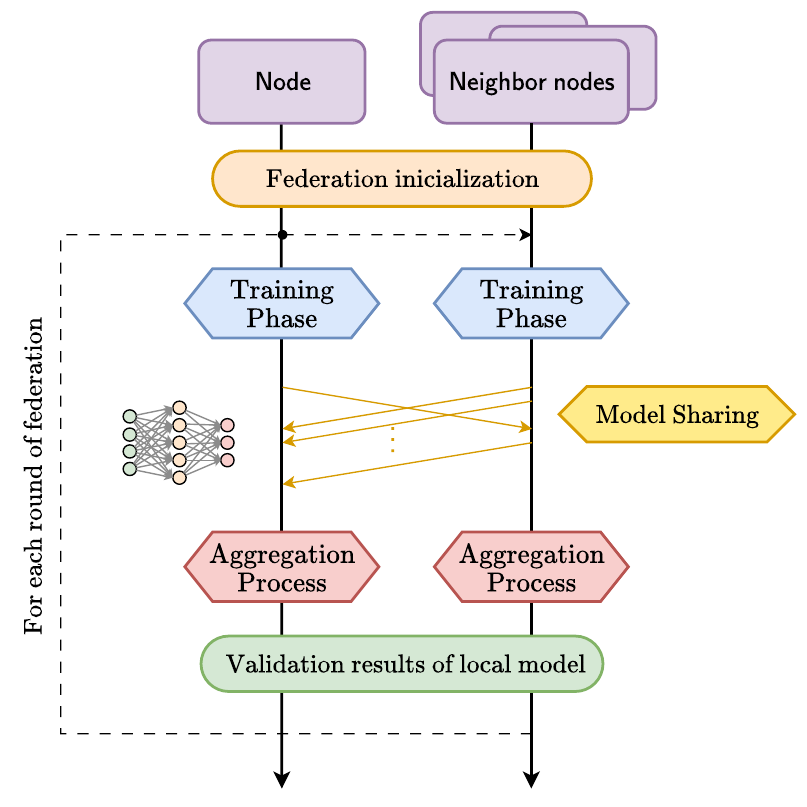}
    \caption{Flowchart of the DFL process in the learning scenario}
    \label{fig:flowchart_dfl}
\end{figure}

In each round of the federation process (hereinafter the terms round and federation round are used interchangeably), every node performs the following steps (summarized in Figure~\ref{fig:flowchart_dfl}): (i) trains its local model using the data available in that round, i.e., batch-based learning; (ii) shares its trained model parameters with peer nodes; and (iii) aggregates the received model parameters using a predefined aggregation algorithm. 

\textbf{Batch-based continuous training:} Traditional FL typically assumes that each node trains on a static and pre-collected local dataset. In contrast, in a real-world scenario, network traffic arrives continuously, and the local dataset is constantly updated. To reflect this, our continuous FL approach processes incoming DoH flows in sequential batches. Each node uses the flows received between consecutive rounds as a new batch to update its local model. This mechanism enables the framework to remain responsive to evolving malicious DoH activity by updating models in line with current traffic behavior, thereby preventing misalignment with recent network patterns. CO-DEFEND design prioritizes per-round refresh cycles with low overhead while keeping raw traffic samples local for privacy. Furthermore, the use of tree-based models such as DT or RF inherently supports learning from stream-wise data with some preservation of stable decision structure under moderate distribution shifts. Finally, the decentralized aggregation itself allows for knowledge reinforcement, as models from different peers (who might retain different aspects of past knowledge) are aggregated.

\textbf{Model parameters sharing:} During each federation round, each entity randomly selects a peer from the P2P network. It shares its trained model parameters using the Gossip communication protocol~\cite{Boyd06}. This protocol is widely known for its efficiency in the rapid dissemination of information over large networks as well as for its convergence capability. One of its main benefits is the reduction of the bandwidth consumption, as it avoids sharing information with all nodes in the network.

%\textbf{Aggregation process:} Various aggregation techniques can be employed within the DFL framework, including statistical averaging~\cite{Yin18}, robust aggregation methods~\cite{Blanchard17}, and ad-hoc methods for non-parametric models such as DT~\cite{Truex19}. The following subsection details the ML algorithms selected for collaborative learning and the model parameter aggregation process for each of them. \textcolor{blue}{Additionally, our proposal places a priority on statistical aggregation and model selection/pruning procedures rather than on security-based techniques. While the latter provide a useful validation-based filter, we intend to leave full Byzantine-robust defenses to future work (see Section~\ref{sec:conclusion})}.

\textbf{Aggregation process:} Various aggregation techniques can be employed within the DFL framework, including statistical averaging~\cite{Yin18}, robust aggregation methods~\cite{Blanchard17}, and ad-hoc methods for non-parametric models such as DT~\cite{Truex19}. The Subsection~\ref{subsec:ml_algorithms} details the ML algorithms selected for collaborative learning and the model parameter aggregation process for each of them. In this work, we prioritize statistical aggregation and model selection/pruning methods over Byzantine-robust defenses, leaving the latter for future work (see Section~\ref{sec:conclusion}). This choice aligns with the considered assumptions defined in the following subsection.

\subsection{Threat model}

The threat model for the CO-DEFEND framework is focused on an external, active adversary whose primary objective is to compromise network security and evade detection. The core of this threat lies in the adversary's use of DoH tunneling to establish covert communication channels for malicious activities, such as C2 operations or data exfiltration. The adversary aims to bypass traditional IDS, making it difficult to distinguish the adversary's malicious flows from the legitimate, benign DoH and HTTPS traffic present in the network.

The CO-DEFEND framework is designed to counter this threat by collaboratively analyzing packet sizes, timing, and flow characteristics, to detect the patterns and anomalies indicative of malicious tunneling. The benign DoH traffic flows collected by each participating entity (e.g., DNS providers, enterprises) is considered highly sensitive as it can reveal user browsing habits and other private information. The system's core design goal is to protect this data from being shared by using DFL. So, CO-DEFEND operates under the assumption that all DFL participants are benign (non-Byzantine), meaning the security focus is strictly on defending against external attacks, not on adversarial actions such as model poisoning or inference attacks from malicious DFL participants.

\subsection{ML Algorithms for Federated Environments}
\label{subsec:ml_algorithms}
Based on previous studies~\cite{Corea24}, this work considers the following machine learning algorithms: SVM, LR, DT, and RF. The selection of these classical algorithms over more computationally intensive models, such as deep neural networks (DNNs), was a deliberate decision. While DNNs are prominent in FL research, their training and communication overhead can be substantial, particularly in decentralized settings with potentially resource-constrained participants. Our choice was guided by several factors, as the aforementioned ML algorithms: (i)  generally exhibit lower computational demands for both training and inference compared to DNNs, making them more amenable to the diverse capabilities of entities in a DFL network; (ii) typically have more compact parameter representations (e.g., weight vectors or tree structures) than large DNNs, leading to reduced bandwidth requirements for model exchange in FL; (iii) are well-suited for continuous and batch-based learning, particularly with adaptations like Stochastic Gradient Descent for SVM/LR and the use of Hoeffding Trees for DT/RF; and (iv) offer greater insight into decision-making processes (especially for DTs)  compared to the black-box nature of many DNNs, which is valuable in security applications and aligns with the Explainable AI (xAI) concept.
We define the ML models, focusing on the description of their adaptation for federated settings. The adaptations required for the continuous learning approach of ML algorithms are described in detail in Section~\ref{sec:implementation}.

\subsubsection{Support Vector Machine}
SVM is a supervised ML algorithm primarily used for classification tasks~\cite{Cortes95}. Its main objective is to find the optimal hyperplane 
\begin{align}
    \label{eq:svm}
    \mathcal{S} &= \left\{ \mathbf{x} \in \mathbb{R}^n \ : \ \mathbf{w}^T \mathbf{x} + b = 0 \right\}, \\
    \mathbf{w}&=\left( \alpha_1,\dots,\alpha_n \right)^T, \nonumber
\end{align}
that maximizes the margin between two classes in a $n$-multidimensional feature space. This margin is defined as the Euclidean distance between the hyperplane and the nearest data points from each class, known as support vectors.

In binary classification, SVM focuses on finding a single hyperplane, which can be characterized by its model parameters $(\mathbf{w},b)$ (weight vector and bias term, both of them are defined in Equation~\ref{eq:svm}), depending on the number of features $n$. Therefore, in a FL scenario, entities only share these parameters with their participants. For the aggregation process, each node (or a central server, if applicable) applies statistical methods to derive a single hyperplane from local updates, typically by computing the coordinate-wise mean of the received weight vectors and bias terms. For instance, if a node $i$ with model $(\mathbf{w}_i, b_i)$ receives a set of models $\mathcal{W}_i=\{(\mathbf{w}_j, b_j)\}_{j \neq i}$ from different peers in the learning network, it can update its local model by averaging $\mathbf{w}_i' = (\mathbf{w}_i + \sum_{j\neq i} \mathbf{w}_j)/(1+|\mathcal{W}_i|)$ and $b_i' = (b_i + \sum_{j\neq i} b_j)/(1+|\mathcal{W}_i|)$. This is a form of Federated Averaging~\cite{McMahan17} applied locally by each DFL participant. This is a form of Federated Averaging~\cite{McMahan17} applied locally by each DFL participant.

\subsubsection{Logistic Regression}

LR is also a supervised ML algorithm primarily used for binary classification, though it can be extended to multi-class problems~\cite{Wright95}. It is based on maximum likelihood estimation (MLE), modeling the probability that a given input belongs to a particular class. It estimates the parameters of the logistic function (also called the sigmoid function):
\begin{align}
    \mathbb{P}(y=1 \mid \mathbf{x}) = \sigma(\mathbf{z}) = \dfrac{1}{1+\exp\left( -\mathbf{z} \right)}, \quad \mathbf{z} = \mathbf{w}^T \mathbf{x} + b,
\end{align}
where $\mathbf{z}$ is a linear representation of the features depending on the coefficients $\mathbf{w}=\left( \beta_1,\dots,\beta_n \right)$, in order to maximize the likelihood of observing the given data.

In order to model the probability of one class relative to the other in a collaborative manner, the shared model parameters are the feature coefficients and the bias term $\left(\mathbf{w},b\right)$. Aggregation methods, such the averaging method proposed for SVM algorithm, are used to combine the parameters from different entities.

\subsubsection{Decision Trees}
\label{subsec:dt}
DTs are non-parametric supervised learning models that recursively partition the feature space through axis-aligned splits by selecting feature thresholds that maximize a node purity metric (e.g., information gain or Gini impurity)~\cite{Breiman84, Quinlan86}. We employ the Hoeffding Tree algorithm as described in~\cite{Domingos00} to support online, batch-based learning from streaming data. Hoeffding Tree is an incremental, anytime decision tree induction algorithm that is capable of learning from massive data streams, assuming that the distribution generating examples does not change over time. It exploits the fact that a small sample can often be enough to choose an optimal splitting attribute. Hoeffding Trees guarantee asymptotic equivalence to batch-trained trees under the Hoeffding bound
\begin{align}
    \epsilon &= \sqrt{\frac{R^2\log(1/\delta)}{2N}},
\end{align}
where $R$ is the range of observer values, $\delta$ represents the split confidence parameter and $N$ the number of observed samples at a node. Each tree $T$ maintains a set of splitting rules $\{\phi_1,\dots,\phi_m\}$ where
\begin{align}
    \phi_i = (f_i, \tau_i, \mathcal{L}_i, \mathcal{R}_i),
\end{align}
with $f_j$ denoting the splitting feature, $\tau_j$ the threshold value, and $\mathcal{L}_j$, $\mathcal{R}_j$ the left/right child nodes. Each node (representing a distinct entity) trains its local Hoeffding Tree on its respective data stream, leading to heterogeneous tree structures in terms of depth, split criteria, and leaf distributions. 

Given the structural non-isomorphism of locally trained trees, we eschew parameter-level aggregation and adopt peer-wise model selection. Previous works have explored methods for aggregating decision trees in FL, such as model selection or decision rule voting~\cite{Truex19, Wu20}. In CO-DEFEND, we adopt a model selection strategy based on local validation performance. Specifically, after a node $i$ shares its local Hoeffding Tree $T_i$ and receives a tree $T_j$ from a peer (or several peers, in a general DFL scenario), it performs the following peer-wise model selection steps:
\begin{enumerate}
    \item \textbf{Pool Formation:} Node $i$ creates a pool of candidate trees, $\mathcal{T}_{\text{candidates}}$, comprising its own locally trained tree $T_i$ and all trees received from its peers.
    \item \textbf{Local Evaluation:} Each candidate tree $T \in \mathcal{T}_{\text{cand}}$ is evaluated on node $i$'s dedicated local validation set (which is distinct from its training data and the global validation set used for overall system evaluation), using prediction accuracy as the measure. The accuracy for a candidate tree $T$ over node $i$'s local validation set $\mathcal{V}_i=\{(\mathbf{x}_k, y_k)\}_{k=1}^{N_\text{val}}$ of size $N_{\text{val}}$ is defined as:
    \begin{align}
        \label{eq:accuracy_dt} 
        \mathsf{Acc}(T,\mathcal{V}_i)=\frac{1}{|\mathcal{V}_i|} \sum_{(\mathbf{x},y) \in \mathcal{V}_i} \mathbb{I}\{T(\mathbf{x})= y\},
    \end{align}
    where \(\mathbf{x}\) and \(y\) denote the validation features and labels of a sample from $\mathcal{V}_i$, respectively. Here, \(T(\cdot)\) represents the prediction function of the candidate tree, and \(\mathbb{I}\{\cdot\}\) is the indicator function. 
    \item \textbf{Model Selection:} Node $i$ selects the tree $T_i^{\text{new}}$ from $\mathcal{T}_{\text{cand}}$ that achieves the highest accuracy on its local validation set. This selected tree $T_i^{\text{new}}$ becomes node $i$'s updated model for the subsequent round.
\end{enumerate}
This ensures that each node adopts the tree structure that demonstrates the best performance on its own current view of the data characteristics, promoting the propagation of effective decision logic across the decentralized network. This policy is computationally lightweight and communication-efficient for non-IID cross-silo DFL; scoring uses accuracy secondary on a  local validation split, with ties broken by smaller depth.

\begin{algorithm}[H]
\caption{DT peer-wise model selection (per round at node $i$)}
\begin{algorithmic}[1]
\Require local tree $T_i$, received trees $\mathcal{T}_\text{cand}$, local validation set $\mathcal{V}_i$
\State $\mathcal{C} \leftarrow \{T_i\} \cup \mathcal{T}_\text{cand}$
\For{$T \in \mathcal{C}$}
  \State Compute $a_T \leftarrow \mathsf{Acc}(T,\mathcal{V}_i)$
  \State Compute $d_T \leftarrow \mathsf{Depth}(T)$
\EndFor
\State $\mathcal{C}_{\text{sorted}} \leftarrow \mathsf{Sort}(\mathcal{C}) \text{ using keys } \langle a_T \downarrow, d_T \uparrow \rangle$
\State Update local tree: $T_i \leftarrow \mathsf{Head}(\mathcal{C}_{\text{sorted}})$
\end{algorithmic}
\end{algorithm}

\subsubsection{Random Forest}
\label{subsec:rf}
RF~\cite{Breiman01} is an ensemble learning method that constructs multiple decision trees and aggregates their predictions to improve overall accuracy and robustness. It builds an ensemble of decision trees by training each on a bootstrapped sample of the data and employing random feature selection at each split. For an input $\mathbf{x} \in \mathbb{R}^n$, the ensemble prediction is determined by majority vote:
\begin{align}
    \hat{y} = \mathsf{mode}\left\{T_1(\mathbf{x}), T_2(\mathbf{x}), \dots, T_N(\mathbf{x})\right\},
\end{align}
To enable collaborative training associated with a DFL scenario, we implement an merge and pruning strategy, similar to approaches studied in federated random forest models~\cite{Markovic22, Souza20}. Specifically, at each step, a node $i$ first performs an ensemble merging step by combining all individual Hoeffding trees from its current local RF ensemble with all individual trees received from the RF ensemble(s) of its peer(s), creating a temporary, larger ``flattened'' pool of trees. Following this, each tree within this merged pool is individually evaluated for its prediction accuracy on node $i$'s dedicated local validation set, using the same accuracy metric as defined in Eq.~\ref{eq:accuracy_dt}. Finally, the forest is pruned: node $i$ selects the top-$N_{\text{target}}$ performing trees from the merged pool, where $N_{\text{target}}$ is the desired size of its RF ensemble (e.g., its original ensemble size). These selected trees form the new, updated RF ensemble for node $i$. This strategy ensures that the aggregated model retains high predictive performance by favoring the most accurate individual trees contributed by the collaboration, helps generalize across varied data distributions inherent to each node, and manages the ensemble size. This pruning strategy aligns with findings in previous federated RF research, which indicate that selectively retaining the most generalizable trees enhances robustness and prevents model overfitting~\cite{Markovic22, Wu20}. \\

\begin{algorithm}[H]
\caption{RF ensemble merge-and-prune (per round at node $i$)}
\begin{algorithmic}[1]
\Require Local forest $F_i$, received forests $\mathcal{F}$, local validation set $\mathcal{V}_i$, target size $N_{\text{target}}$
\State $\mathcal{P} \leftarrow F_i \cup \left( \bigcup_{F \in \mathcal{F}} F \right)$
\For{$T \in \mathcal{P}$}
  \State Compute $a_T \leftarrow \mathsf{Acc}(T,\mathcal{V}_i)$
  \State Compute $d_T \leftarrow \mathsf{Depth}(T)$
\EndFor
\State $\mathcal{P}_{\text{sorted}} \leftarrow \mathsf{Sort}(\mathcal{P}) \text{ using keys } \langle a_T \downarrow, d_T \uparrow \rangle$
\State Update local forest: $F_i \leftarrow \{ T \in \mathcal{P}_{\text{sorted}} \, : \, \mathsf{Rank}(T) \leq N_{\text{target}} \}$
\end{algorithmic}
\end{algorithm}

\textbf{Why selection instead of aggregation for trees?} Unlike linear models, decision trees do not share an aligned parameter space across peers; node tests $(f,\tau)$ and subtrees are learned under different local marginals. Averaging or na\"ive subtree fusion breaks impurity objectives (Gini/IG) and often yields invalid split tests. Moreover, aligning heterogeneous trees requires tree-edit/matching procedures whose computational cost scales poorly with depth/size and imposes significant communication overhead in DFL. Consequently, \textsc{CO-DEFEND} adopts peer-wise model selection (DT) and ensemble merge-and-prune (RF) driven by each peer's held-out, local validation: a node adopts the best-scoring received tree (DT) or retains the top-$N_{\text{target}}$ trees after merging remote ensembles (RF). This selection pressure propagates generalizable decision logic under different constraints while keeping communication lightweight. This design aligns with how federated tree systems avoid parameter averaging in practice (histogram/protocol-based training or ensemble-level strategies.

\subsection{Convergence Analysis of CO-DEFEND}
\label{subsec:convergence_analysis}

We establish a clear convergence rate guarantee for the proposed algorithm (collaborative training in gossip-based DFL settings), demonstrating that it achieves a stationary-point optimal solution. Convergence in decentralized learning frameworks has been widely examined in prior works~\cite{yang2022decentralized, yuan2016convergence, xin2021improved}. Our analysis for the non-convex case closely follow seminal convergence proof in~\cite{liu2022decentralized}, which is particularly well suited to the problem under study. The analysis builds on standard assumptions in the field, such as Lipschitz continuity of gradients, unbiased stochastic gradients, and bounded gradient variance.

In the framework considered in~\cite{liu2022decentralized}, the communication graph is static, and each participant exchanges its local model only with direct neighbors. In contrast, our gossip-based setup assumes a fully connected communication graph, enabling every node to interact with all others, a reasonable setting for cross-silo federated scenarios. Furthermore, our method incorporates a gossip-based communication mechanism, where in each round, every node randomly selects peers to exchange models with, according to a specific probability. This random pairwise exchange pattern effectively induces an Erdős–Rényi random graph at each round, whose spectral properties determine how fast information spreads across the network. Formally, the mixing matrix $W \in \mathbb{R}_{|\mathcal{V}| \times |\mathcal{V}|}$ represents the weighted adjacency of the peer-to-peer communication graph, satisfying $W=W^\top$ and $W \cdot \mathbf{1} = \mathbf{1}$. Also, the spectral gap, defined as $1-\zeta$ with $\zeta =\lambda_2(W)$ being the second-largest eigenvalue of $W$, measures the rate of information diffusion, i.e., a larger spectral gap implies faster consensus among nodes, while a smaller gap slows information diffusion. The spectral gap of Erdős–Rényi graphs has been extensively studied in the literature~\cite{hoffman2021spectral, feige2005spectral}.

Following the derivation in~\cite{liu2022decentralized}, the convergence rate in the non-convex DFL case can be upper bounded by $\mathcal{O} \left( 1/T \right) + \mathcal{O} \left( 1/|\mathcal{V}| \right) + \mathcal{O} \left( \zeta^2/(1-\zeta^2) \right)$, where $T$ denotes the number of federation rounds. This expression is essentially a reformulation of~\cite[Eq. 20]{liu2022decentralized} in big-$\mathcal{O}$ notation, after removing lower-order constants and adapting the update and communication frequencies to our setup. The first two terms stem from the stochastic gradient descent approximation error, while the third term accounts for the consensus error due to local model drift. As expected, the SGD-related error decreases with more nodes and training rounds, since averaging across a larger set reduces variance. By contrast, the consensus error is not directly influent by the number of nodes; it depends primarily on the spectral properties of the mixing matrix, with sparser connectivity leading to slower convergence.

In summary, this analysis confirms that the gossip-based DFL setting under consideration falls within the scope of established non-convex convergence guarantees. Despite the introduction of probabilistic communication and random graph structures, the convergence rate aligns the same asymptotic behavior as in~\cite{liu2022decentralized}. This allows us to rigorously argue that, even under our gossip-based communication model with probabilistic exchanges, decentralized training preserves the theoretical guarantees and converges toward a stationary-point optimal solution.

\section{EXPERIMENTAL SETUP}
\label{sec:implementation}

In this section, we detail the experimental setup and the various scenarios used to evaluate the proposed DFL framework. We also describe the considered dataset for validation and the software implementation used for the simulations.

\subsection{Dataset}
\label{subsec:dataset}

The dataset used in this study is CIRA-CIC-DoHBrw-2020~\cite{montazeri2020detection}, which was collected using a purpose-built traffic-capturing infrastructure designed to generate both benign and malicious DoH traffic. The corresponding benign traffic was generated by browsing the top 10,000 Alexa-ranked websites using Google Chrome and Mozilla Firefox, both supporting DoH. In contrast, malicious DoH traffic is generated using DNS tunneling tools such as dns2tcp~\cite{dns2tcp}, DNSCat2~\cite{dnscat2}, and Iodine~\cite{iodine}, which encapsulate data within DNS queries to create encrypted data tunnels. To ensure a comprehensive evaluation of DoH servers, authors used the following four providers: AdGuard, Cloudflare, Google, and Quad9. 

The dataset consists of statistical features derived from the captured traffic, which were extracted using a tool called DoHLyzer, developed by the authors. The proposed tool groups DoH packets into flows, where a DoH flow is defined as a sequence of one or more packets that have the same source and destination. The distinction of a DoH flow is the time constraint between packets: there is a defined maximum time interval that can elapse between two packets within a flow. DoHLyzer extracts 28 statistical and time-series features from the captured traffic and outputs them, along with flow duration, IP addresses, ports, and initial timestamp, in a CSV file, as detailed in~\cite{montazeri2020detection}. Additionally, it also includes a label indicating whether the traffic is benign or malicious. As a result, the dataset comprises 34 features along with the corresponding label.

Our study exclusively utilizes the CIRA-CIC-DoHBrw-2020 dataset because, based on a comprehensive review of the literature, it is the only large-scale, publicly available benchmark specifically designed for DoH tunneling detection. Its unique value lies in providing explicitly labeled malicious flows generated by well-known tunneling tools (dns2tcp, dnscat2, Iodine), mixed with benign DoH traffic. Even recent studies aiming to address this dataset's limitations~\cite{niktabe2024unveiling}, have focused on creating balanced subsets from the original CIRA-CIC-DoHBrw-2020 data rather than introducing a new, independent corpus. Consequently, its use is standard practice for ensuring reproducibility and enabling direct comparison with state-of-the-art approaches.

\textbf{Dataset Distribution:} To evaluate a federated environment, we aim to simulate DNS traffic from different entities. While this could be achieved by randomly partitioning the dataset, as done by Li et al. in~\cite{li2022detecting}, we adopt an alternative approach to ensure that each entity's traffic exhibits distinct characteristics. Specifically, we simulate a scenario in which each entity utilizes a different DNS provider, ensuring a more realistic scenario where traffic distribution varies between entities. We segment the traffic by DNS provider based on IP addresses, enabling evaluation per provider. Table~\ref{tab:dataframe_sizes} summarizes the flow distribution across providers. As can be observed, the traffic is unbalanced and the number of malicious flows is significantly higher than the number of benign flows. 

\begin{table}
    \centering
    \caption{Number of flows and their malicious/benign distribution}
    \label{tab:dataframe_sizes}
    \setlength{\tabcolsep}{3pt}
    \begin{tabular}{|c|c|c|c|c|}
        \hline
        \textbf{Entity} & \textbf{DNS Provider} & \textbf{Total Size} & \textbf{Malicious} & \textbf{Benign}\\ 
        \hline
        1 & GoogleDNS & 53,586 & 41,574 & 12,012 \\ 
        2 & Cloudflare & 31,226 & 29,346 & 1,880 \\ 
        3 & AdGuard & 25,144 & 22,224 & 2,920 \\ 
        4 & Quad9 & 159,687 & 156,692 & 2,995 \\ 
        \hline
    \end{tabular}
\end{table}

\textbf{Training/Validation segmentation:} Prior to distributing the dataset to each entity, we selected a random $10\%$ of the total samples as a global validation set containing benign samples and attack patterns belonging to each entity. This validation set is used by all entities (or the central server, if applicable) to evaluate each of the local models under the same conditions and observe whether the local models generalize for different attack patterns. This criterion is used for system validation purposes only. The remaining dataset is divided by IP addresses between each of the entities as described above. In each of these entities, another $10\%$ of the samples of the local dataset is selected for a local validation of the own models and to check if they better detect attacks performed on the entity itself. Finally, the remaining samples are used to train each local model of the entities and are packaged in batches for training in each of the federation rounds. The proportion of benign/malicious samples in each of the batches is in accordance with the proportion shown in each entity's trainset.

\textbf{Feature Selection:} PCA is used in this study to reduce the dimensionality of the feature space and address challenges associated with high-dimensional data. By reducing the number of features, PCA not only improves computational efficiency but also enhances model performance by focusing on the most informative aspects of the data while eliminating redundant or less relevant features. Moreover, PCA helps mitigate the risk of overfitting, as it simplifies the model without compromising its ability to generalize. 

\textbf{Privacy measure:} The use of PCAs can also contribute to privacy preservation, reducing the risk of pattern imitation by adversaries, which strengthens the model against evasion techniques. As shown in previous studies~\cite{moure2023real}, there are certain features that provide sufficient information to identify patterns in malicious traffic. However, attackers may mimic these features to avoid detection. The projection into an orthogonal principal-subspace can also raise the bar for simple feature-level mimicry, as an attacker now needs to reproduce the joint covariance structure of the benign flows rather than imitate a small set of raw features. As previously established in works such as~\cite{bhagoji2018enhancing,alemany2020dilemma}, data transformations (including PCA) can be used as defenses against evasion attacks when inputs are projected to a lower-dimensional subspace. We therefore position PCA as a complementary privacy measure used here together with FL techniques.

\subsection{Definition of Validation Scenarios}
\label{subsec:scenarios}

\begin{figure*}
    \centering
    \includegraphics[width=\textwidth]{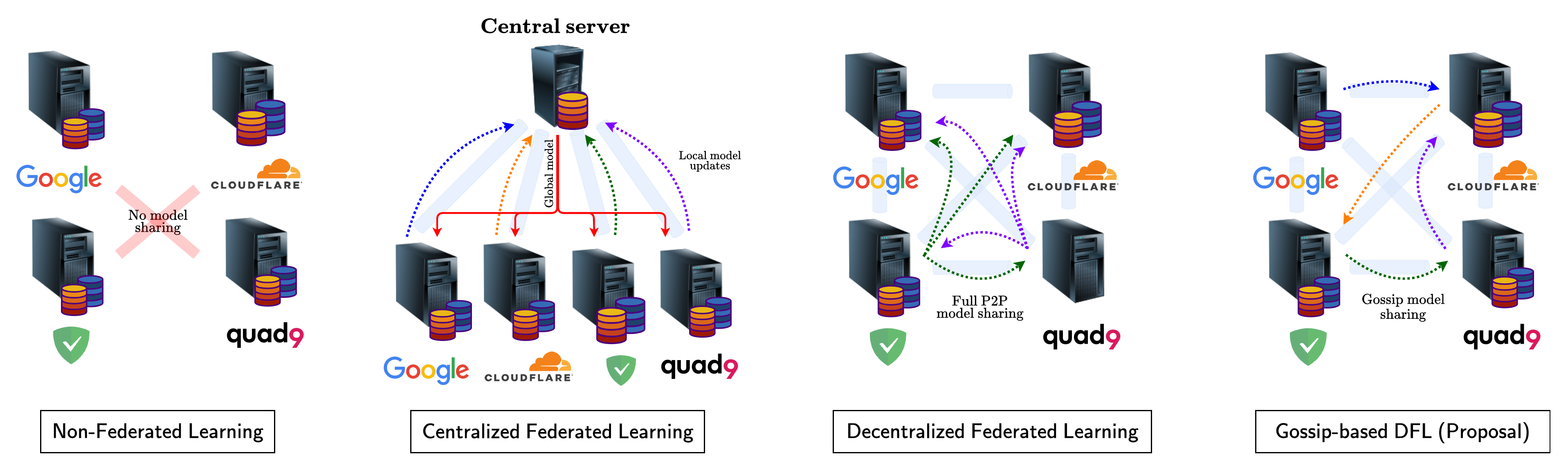}
    \caption{Description of the four evaluation scenarios for collaborative training between entities with different DNS providers: non-federated learning, centralized federated learning, decentralized federated learning and Gossip-based decentralized federated learning. In all scenarios, both nodes and central server, the ML model validation sets contain the same samples.}
    \label{fig:scenario_evaluation}
\end{figure*}

To evaluate the effectiveness of our approach, we consider four different training scenarios (summarized in Figure~\ref{fig:scenario_evaluation}) that cover both centralized and decentralized learning paradigms:

\begin{itemize}
    \item \textbf{Non-Federated Learning (Baseline):} Each node trains a local model using only its own local data, with no collaboration between nodes, i.e. no model parameters are exchanged in this scenario. This represents a conventional centralized approach \textit{per node} (no federation) and serves as a baseline to see how well an individual node can detect attacks in isolation.
    
    \item \textbf{Centralized Federated Learning:} Each node participates in a standard FL setup with a central server (aggregator). In each round, nodes train local models on their respective data and send the model parameters to the central server. The server aggregates these models (e.g., by averaging parameters for SVM/LR or selecting decision trees as described in Section~\ref{sec:methodology}) into a single global model, which is then broadcast back to all nodes for the next round. This scenario is equivalent to the classical CFL framework and allows us to compare decentralized vs. centralized aggregation.
    
    \item \textbf{Fully-connected Decentralized Federated Learning:} The nodes collaborate in a P2P network without any central coordinator. Each node communicates directly with others to share model updates. In our implementation, during each round every node exchanges model parameters with all other peers (fully connected topology). Each node then aggregates updates from peers locally. This yields a global consensus model over time, but achieved via distributed agreement. The DFL scenario tests robustness to the removal of the central aggregator, potentially improving fault tolerance and decentralization at the cost of more complex communication.

    \item \textbf{Gossip-based Decentralized Federated Learning (aligns with proposal):} A modified decentralized scenario where model sharing is based on a randomized Gossip protocol. In each round, instead of broadcasting to all peers, each node randomly selects one peer and exchanges its model with that peer. These pairwise exchanges are randomized each round, and over many rounds the updates propagate through the network (Gossip ensures eventual consistency). This approach significantly reduces communication overhead per round while still allowing the federation to reach a common model. The Gossip-based DFL scenario is more scalable to large networks.
\end{itemize}
All scenarios are evaluated under identical experimental conditions, including the same data partitioning, learning models, and simulation parameters. The FL process is conducted over $20$ rounds. The batch size in all entities is $153$ samples. This is the number of samples available to AdGuard (the provider with the lowest number of samples) in each of the rounds if the train set allocation is equal in each round. Due to the disparity in the number of samples between entities, this allocation of samples in batches is done to ensure a fair comparison in each of the rounds.

By comparing these scenarios, we can assess how a fully centralized approach, a conventional federated approach (CFL), and our decentralized approaches (DFL with full sharing vs. DFL with gossip) perform relative to each other in detecting malicious DoH tunnels.

\subsection{Performance Metrics}

To measure the detection performance of each model, we use four standard classification metrics: accuracy, precision, recall, and F1-score ~\cite{hossin2015review}. These metrics are computed from the counts of true positives, false positives, true negatives, and false negatives obtained on the test data. We briefly define each metric below (TP = True Positives, FP = False Positives, TN = True Negatives, FN = False Negatives):

\begin{itemize}
    \item \textbf{Accuracy} is the fraction of all queries that are correctly classified:
    \begin{equation}
        A = \frac{\text{TP} + \text{TN}}{\text{TP} + \text{TN} + \text{FP} + \text{FN}}
    \end{equation}

    \item \textbf{Precision} (Positive Predictive Value) is the proportion of predicted malicious queries that are truly malicious:
    \begin{equation}
        P =\frac{\text{TP}}{\text{TP} + \text{FP}}
    \end{equation}

    \item \textbf{Recall} (True Positive Rate) is the proportion of actual malicious queries that are correctly detected:
    \begin{equation}
        R=\frac{\text{TP}}{\text{TP} + \text{FN}}
    \end{equation}

    \item \textbf{F1-score} is the harmonic mean of precision and recall:
    \begin{equation}
        F1=\frac{2 \times \text{TP}}{2 \times \text{TP} + \text{FP} + \text{FN}} = 2 \times \frac{P \times R}{P + R}
    \end{equation}
\end{itemize}

These metrics provide a comprehensive evaluation of model performance. Accuracy gives an overall correctness rate, precision reflects how reliable the model’s positive (malicious) predictions are (low false-alarm rate if precision is high), recall indicates the model’s ability to catch malicious instances (coverage of true attacks), and F1 balances the two, which is useful if there is an imbalance between benign and malicious classes. In the context of our problem, precision corresponds to how many of the detected malicious DoH flows were actually malicious (for avoiding false alarms), while recall corresponds to how many of the actual malicious flows we managed to detect (for security coverage).

\subsection{Heterogeneity Analysis Metrics}

In this subsection, we introduce the metrics used to quantify the statistical distance between provider datasets and conduct a comparative statistical analysis to highlight their similarities and differences. To formally measure the non-IID nature of the data distribution across providers, we employ a suite of statistical divergence metrics. First, to ensure feature divergences are comparable across dimensions, we globally standardize each numeric feature \(f\) using its mean \((\mu_f)\) and standard deviation \((\sigma_f)\) computed over the entire dataset:
\begin{equation*}
  x_f \leftarrow \frac{x_f - \mu_f}{\sigma_f}  
\end{equation*}
Let \(p\) and \(q\) index two distinct provider partitions.
\begin{itemize}
\item \textbf{Label Skew:} We measure the difference in class distributions using the Total Variation (TV) distance, which for a binary classification task simplifies to the absolute difference in the fraction of malicious samples:
\begin{equation}
D_{\text{lbl}}(p,q) = \bigl| \mathbb{P}_p(\text{malicious}) - \mathbb{P}_q(\text{malicious}) \bigr|.
\end{equation}

\item \textbf{Feature Divergences:} For each standardized feature \(f\), we compute two complementary well-known divergence measures: the scale-invariant \emph{Wasserstein-1 Distance} \(\widehat{W}_{1,f}\) and the bounded \emph{Jensen--Shannon Divergence} \(\mathrm{JS}_f\).

\item \textbf{Non-IID Index (NII):} We aggregate these metrics into a single pairwise \emph{Non-IID Index (NII)} to provide a holistic measure of heterogeneity:
\begin{equation}
\mathrm{NII}(p, q) = \alpha\, D_{\text{lbl}}(p, q) + (1-\alpha)\, \sum_{f=1}^{d} w_f \bigl( \lambda\, \widehat{W}_{1,f} + (1-\lambda)\, \mathrm{JS}_f \bigr).
\end{equation}
\end{itemize}

By default, we use uniform feature weights \((w_f = 1/d)\), \(\alpha = 0.3\), and \(\lambda = 0.5\).

\subsection{Software Implementation}

A distributed computing platform is required to deploy the DFL process. After reviewing state-of-the-art simulation tools~\cite{Beltran23,Riedel24}, we selected the open-source simulator\footnote{Code is available online: \\ \textsf{\url{https://gitlab.com/compromise3/decentralizedfedsim}}} developed in~\cite{cajaraville2024byzantine}. It is an open-source simulator developed in Python for the execution of tests in the DFL framework, although it allows configuring CFL scenarios and distributed non-collaborative scenarios. Furthermore, this simulator has been designed to be scalable, allowing the inclusion of ad-hoc code for training algorithms or distributed dataset processing without requiring a focus on the DFL process itself. The adaptation of the simulator for the validation of our methodology is also open-source\footnote{Code is available online: \\ \textsf{\url{https://gitlab.com/compromise3/co-defend}}}, and the specific changes made are specified below.

For this purpose, several classes/functions have been developed to implement data collection, preprocessing, and distribution, as detailed in this section, as well as the ML models chosen for the classification of traffic flows.

For the implementation of these models, the Python library \texttt{sklearn} has been used, which provides the necessary tools for the SVM and LR algorithms, including support for continuous training of the samples through the means of the \texttt{SGDClassifier} class. The RF and DT algorithms have been adapted from the \texttt{River} library, which provides the underlying tree structures and their processing, although it does not natively support continuous training. For the trees in both RF and LR models, the \texttt{HoeffdingTreeClassifier} class is employed. Additionally, the aggregation algorithms required for the model sharing process, as detailed in Section~\ref{sec:methodology}, have also been implemented. 

\section{EVALUATION AND DISCUSSION}
\label{sec:evaluation}

In this section, we will present the results for each scenario and discuss how the models fared according to these metrics. Prior to the presentation of the results, a feature selection procedure is performed by using PCA, motivated by the challenges posed by high-dimensionality, the risk of overfitting, and privacy concerns in the federated environment (Subsection~\ref{subsec:pca}). Furthermore, a heterogeneity analysis of the dataset is conducted to assess the impact of collaborative learning on the performance of machine learning models in our case study (Subsection~\ref{subsec:heterogeinity_analysis}). Following these previous study, we evaluate the performance of the various algorithms (as defined in Section~\ref{sec:methodology}) across the scenarios described in Section~\ref{sec:implementation}. In this way, the results are discussed under the following criteria: (i) comparison between the validation scenarios (Subsection~\ref{subsec:scenarios_comparison}), and (ii) comparison between the trained algorithms and other similar works (Subsection~\ref{subsec:ml_comparison}). Finally, we assess a scalability analysis under the DFL+Gossip scenario to evaluate how performance is affected by the number of nodes and the model sharing frequency (Subsection~\ref{subsec:scalability_analysis}).

\subsection{Feature Selection -- PCA}
\label{subsec:pca}

\begin{figure*}
    \centering
    \includegraphics[width=\textwidth]{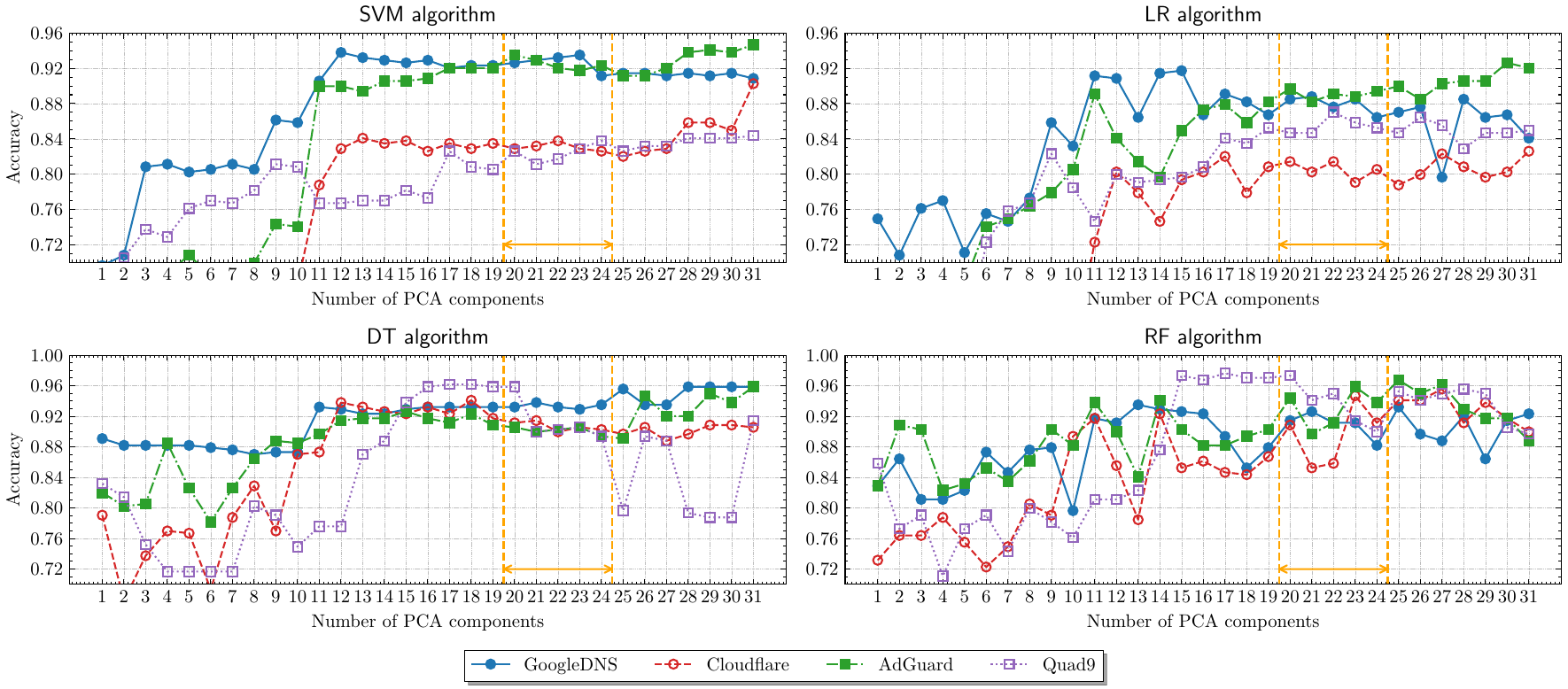}
    \caption{Sweep of the number of PCA components of the accuracy in a full-DFL scenario with 10 rounds}
    \label{fig:pca_sweeping}
\end{figure*}

A key aspect here is the choice of the number of PCA components needed to characterize the dataset without incurring a loss of relevant information. Therefore, we perform a study of the evolution of the accuracy of the different entities in a specific scenario. A full-DFL scheme was defined with 10 federation rounds. We conducted an exhaustive sweep of PCA components, varying the number from 1 up to the full dimensionality of the original dataset, to determine the optimal trade-off between accuracy and feature reduction. The results can be seen in Figure~\ref{fig:pca_sweeping} where each of the plots shows the sweep of the number of PCA components for each of the training algorithms.

In view of the results, selecting an optimal number of PCA components that maximizes the accuracy of all entities in all ML algorithms simultaneously is practically infeasible. Our analysis reveals that performance improves markedly when the number of PCA components exceeds 10, although beyond this threshold, the improvement is inconsistent due to outlier behavior in certain entities (e.g., Quad9 for DT, GoogleDNS for LR). In general, most cases show good results in the range of $20-24$ PCA components (as it is shown).

\begin{figure}
    \centering
    \includegraphics[width=0.5\textwidth]{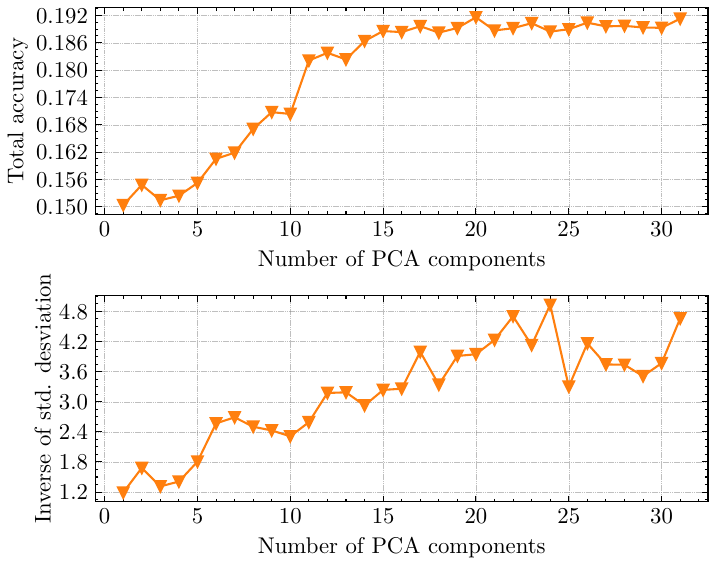}
    \caption{Statistics resulting from the PCA components analysis}
    \label{fig:pca_stats}
\end{figure}

The final selection was decided based on two criteria: (i) the total accuracy of all the scenarios be as high as possible; and (ii) the variability between the results of each entity be as low as possible (avoiding very good results in some entities, and low in others). These normalized results can be seen in Figure~\ref{fig:pca_stats}, where the best results, simultaneously, are found in $22$ and $24$ components. We ultimately selected $22$ PCA components, as this configuration maximized the overall accuracy across scenarios while minimizing inter-entity variability, with only a minor trade-off observed for RF performance. This selection enables a balanced approach by preserving a trade-off between good performance and potential risk of overfitting due to the model's increased complexity. Related to the latter, performance has been thoroughly monitored through cross-validation, with no significant increase in performance beyond $22$ components, in most cases.

\subsection{Heterogeneity Analysis of Dataset}
\label{subsec:heterogeinity_analysis}

\begin{figure}[H]
\centering
\includegraphics[width=0.45\textwidth]{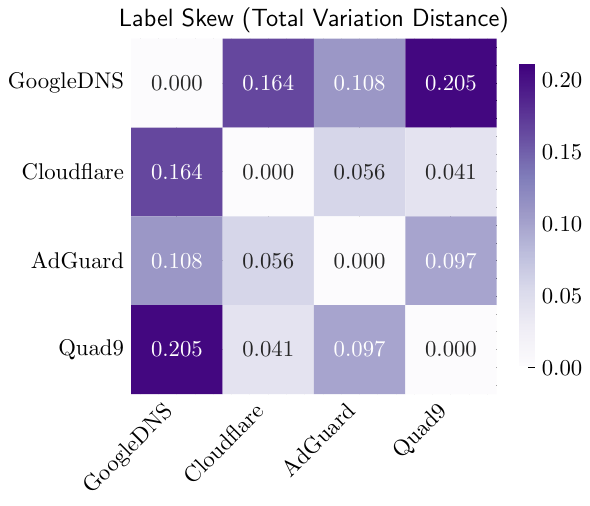}
\caption{Heatmap of pairwise Total Variation distance (Label Skew) across DNS provider datasets. Higher values indicate greater difference in the benign/malicious class ratios.}
\label{fig:label_skew_heatmap}
\end{figure}

These findings provide empirical support for our claims regarding the benefits of the proposed DFL-based CO-DEFEND framework. Our quantitative analysis first confirmed the presence of significant statistical heterogeneity across the DNS-provider-based splitted datasets, validating our methodological design. The degree of imbalance in malicious traffic proportions highlights substantial variations across providers. For example, from 77.6\% of malicious traffic in GoogleDNS samples to 98.1\% in Quad9 samples. This results in a significant pairwise label skew, with a maximum TV distance of 0.205 between GoogleDNS and Quad9, as visualized in Figure~\ref{fig:label_skew_heatmap}. Beyond class imbalance, distributional shifts in traffic features were also evident. Well-known Kruskal--Wallis testing confirmed that numerous features exhibit statistically significant distributional differences across providers. As shown in Table~\ref{tab:top_divergent_features}, where features are ranked by average Wasserstein-1 distance, fundamental traffic characteristics like packet timing and duration are among the most divergent between datasets, underscoring that the datasets differ not just in class balance but in fundamental feature shapes. This combination of label and feature skew illustrates that heterogeneity is both strong and multidimensional.

\begin{table}[H]
\centering
\caption{Top-5 most divergent features, ranked by average pairwise Wasserstein-1 distance. All show highly significant differences across providers.}
\label{tab:top_divergent_features}
\small
\begin{tabular}{|l|c|}
\hline
\textbf{Feature} & \textbf{Avg.\ W1 Distance} \\
\hline
PacketTimeMedian   & 0.914 \\
Duration           & 0.901 \\
PacketTimeMean     & 0.892 \\
PacketLengthMean   & 0.871  \\
PacketTimeVariance & 0.847 \\
\hline
\end{tabular}
\end{table}

\begin{figure}[H]
\centering
\includegraphics[width=0.75\textwidth]{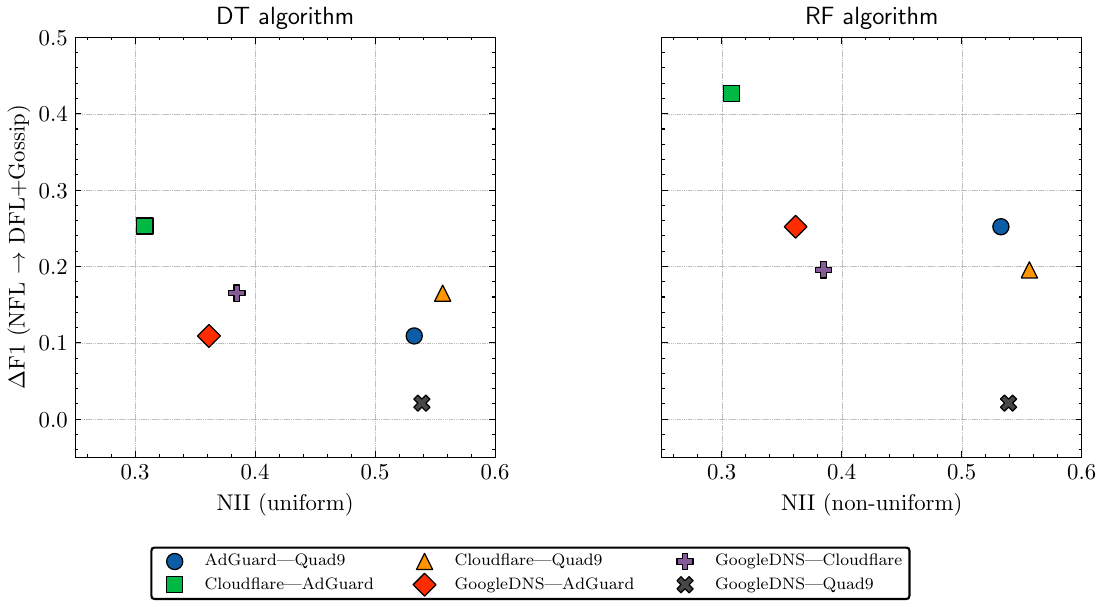}
\caption{Scatter plot of pairwise NII vs.\ average \(\Delta\)F1 for DT and RF models}
\label{fig:corr_scatter}
\end{figure}

Furthermore, we also examined the relationship between the degree of heterogeneity (measured by our Non-IID Index) and collaborative learning gains (measured with variation in F1 scores, $\Delta \mathrm{F1}$). Figure~\ref{fig:corr_scatter} compared the pairwise NII values with the average performance improvements \(\Delta\)F1 for each pair of providers when moving from non-federated learning (NFL) to our DFL+Gossip approach. The results indicate a non-positive correlation (Spearman's rank correlation coefficient \(\rho = -0.765\) for RF), showing that higher heterogeneity does not directly lead to larger gains. Instead, the largest F1-score improvements are observed for providers like Cloudflare (\(\Delta\)F1 \(= +0.370\) for RF) and AdGuard (\(\Delta\)F1 \(= +0.483\) for RF), who had the weakest initial performance results in its local models.

This strong negative correlation suggests that while weaker clients gain substantially from stronger peers, a point of diminishing returns may be reached when provider datasets are excessively dissimilar. In such cases, the models learned by one peer could be too specialized to be effectively adopted by another, highlighting the importance of the local validation step in our selection mechanism to filter out incompatible knowledge.

In summary, these results suggest that the benefits of our approach extend beyond simple data splitting and collaborative learning. In this case, performance gains are primarily driven by knowledge transfer, where our gossip-based selection-and-pruning mechanism allows underperforming clients to effectively learn from and adopt the superior, more generalizable models developed by their peers. This highlights that CO-DEFEND, through DFL techniques and adaption of classical ML models, turns heterogeneity from a potential obstacle into an effective collaboration.

\subsection{Validation scenarios comparison}
\label{subsec:scenarios_comparison}

\begin{figure*}
    \centering
    \includegraphics[width=\textwidth]{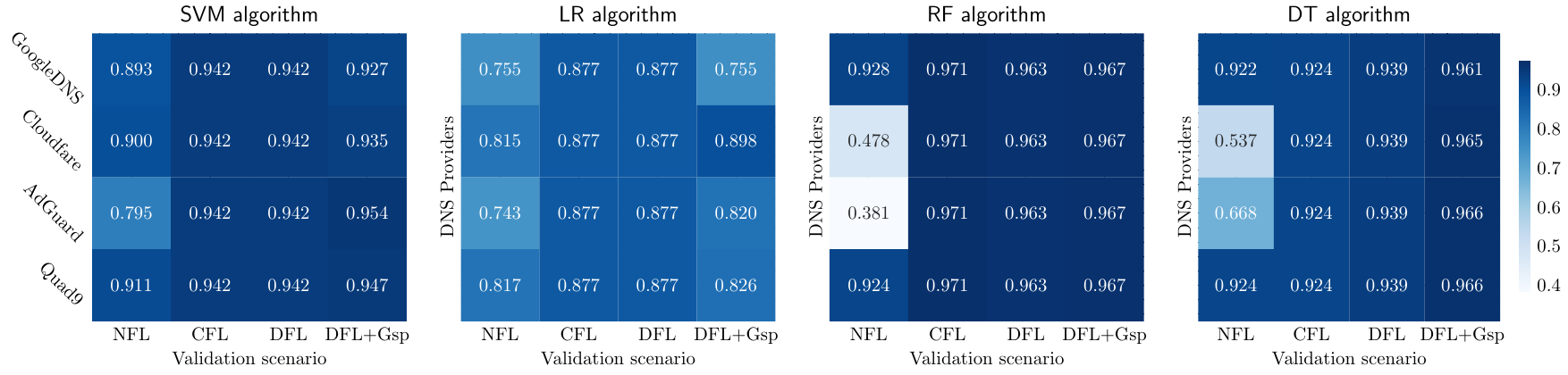}
    \caption{Accuracy for different DNS providers and validation scenarios. NFL: Non-Federated Learning; CFL: Centralized Federated Learning; DFL: Decentralized Federated Learning; DFL+Gsp:  Gossip-based DFL.}
    \label{fig:heatmap_2}
\end{figure*}

Figure~\ref{fig:heatmap_2} presents the final accuracy values achieved by each DNS entity after 20 rounds, illustrating the performance under different validation scenarios and across various ML algorithms. Based on these results, it can be clearly observed how the federation process in any of its forms provides a positive result in all possible scenario-algorithm combinations. Notably, for entities such as GoogleDNS and Cloudflare, tree-based algorithms (DT and RF) exhibit substantial improvements when federation is applied; for example, the RF model’s accuracy more than doubles compared to local training. Using a global testset to validate the models, it is observed how the distributed collaboration process allows the entities to learn about attack patterns that were previously unknown to them (since they did not have such data). 

On the other hand, the SVM and LR algorithms also experience an increase in the results obtained by introducing federation, although it is more discrete (around $3\%$ and $14\%$). This analysis is similar for the AdGuard and Quad9 entities, with the difference that the tree-based algorithms already show significant results. This is predictably due to the nature of tree-based algorithms, improving generalization on unseen attack behaviors. Unlike linear models (SVM, LR), they leverage additional structural patterns. The federated process integrates other participants' decision boundaries, resulting in more accurate classifiers.

As for the federated scenarios, the vast majority of cases show similar results among them, which requires a more refined analysis of the results. In these simulations, the CFL and DFL results are the same in all cases because the aggregated models in each round are identical: in CFL, the central node computes the global model based on all contributions; in DFL, all entities receive the rest of the contributions and compute the same (global) model. However, it is important to note that a fully-connected DFL topology is impractical at scale since its communication cost grows quadratically with the number of nodes, $\mathcal{O}(n^2)$. In contrast, the proposed Gossip-based DFL reduces communication overhead to $\mathcal{O}(n)$, making it more viable for large-scale deployments.

On the other hand, one might expect that the results of DFL+Gossip would be worse than those of the DFL scenario, since the federation scheme is the same, but there is less sharing of ML models between entities. Nothing could be further from the truth, the results in the latter scenario maintain the good metrics obtained in DFL in most cases, and even improve in some entities and algorithms. The network topology is fully connected, allowing each node to randomly select any other node as its gossip partner. This ensures that, over multiple rounds, model updates propagate efficiently throughout the network. This is the case of the DT algorithm, where the proposed scenario obtains the best results among those evaluated. The same case occurs with the RF algorithm, although the improvement in accuracy is almost imperceptible. Others, such as SVM, obtain similar results for practical purposes, with or without the inclusion of the Gossip protocol. As an exception, the LR algorithm is the only ML model evaluated that worsens its results with respect to the full-DFL scenario, predictably due to the lack of convergence in reducing the spread of the models.

\begin{table*}[ht]
    \centering
    \caption{Accuracy for different DNS providers and validation scenarios. NFL: Non-Federated Learning; CFL: Centralized Federated Learning; DFL: Decentralized Federated Learning; DFL+Gsp:  Gossip-based DFL.}
    \label{tab:performance_metrics}
    
    \resizebox{\textwidth}{!}{%
    \begin{tabular}{|c|c|c|c|c|c|c|c|c|c|c|c|c|c|c|c|c|}
        \hline
        \multirow{2}{*}{\textbf{DNS Provider}} & \multicolumn{4}{c|}{\textbf{SVM}} & \multicolumn{4}{c|}{\textbf{LR}} & \multicolumn{4}{c|}{\textbf{RF}} & \multicolumn{4}{c|}{\textbf{DT}} \\ \cline{2-17}
        & \textbf{NFL} & \textbf{CFL} & \textbf{DFL} & \textbf{DFL+Gsp} & \textbf{NFL} & \textbf{CFL} & \textbf{DFL} & \textbf{DFL+Gsp} & \textbf{NFL} & \textbf{CFL} & \textbf{DFL} & \textbf{DFL+Gsp} & \textbf{NFL} & \textbf{CFL} & \textbf{DFL} & \textbf{DFL+Gsp} \\ \hline
        GoogleDNS & 0.893 & 0.942 & 0.942 & 0.927 & 0.755 & 0.877 & 0.877 & 0.755 & 0.928 & 0.971 & 0.963 & 0.967 & 0.922 & 0.924 & 0.939 & 0.961 \\ 
        Cloudflare & 0.900 & 0.942 & 0.942 & 0.935 & 0.815 & 0.877 & 0.877 & 0.898 & 0.478 & 0.971 & 0.963 & 0.967 & 0.537 & 0.924 & 0.939 & 0.965 \\ 
        AdGuard & 0.795 & 0.942 & 0.942 & 0.954 & 0.743 & 0.877 & 0.877 & 0.820 & 0.381 & 0.971 & 0.963 & 0.967 & 0.668 & 0.924 & 0.939 & 0.966 \\ 
        Quad9 & 0.911 & 0.942 & 0.942 & 0.947 & 0.817 & 0.877 & 0.877 & 0.826 & 0.924 & 0.971 & 0.963 & 0.967 & 0.924 & 0.924 & 0.939 & 0.966 \\ \hline
    \end{tabular}
    }
\end{table*}

It is important to consider not only accuracy but also the F1-score, as its significance is critical in scenarios where precision and recall must be balanced to minimize false positives and false negatives. Minimizing false positives is essential to avoid excessive alerts. Misclassifying benign DoH queries as malicious could result in the unnecessary blocking of legitimate services, leading to user access issues and potential service interruptions. 
Table~\ref{tab:performance_metrics_f1} collects the F1-Score values for different DNS providers and scenarios. SVM consistently achieves an F1-score of $0.969$ in both DFL and CFL across all providers, matching centralized performance, while DFL+Gossip slightly improves results, reaching $0.975$ for AdGuard, indicating enhanced model aggregation. Similarly, LR maintains stable performance ($0.93$) across all providers, outperforming centralized learning in some cases, such as GoogleDNS (0.849), with DFL+Gossip further improving generalization, particularly for Cloudflare ($0.945$) and Quad9 ($0.898$).

Tree-based algorithms (RF and DT) exhibit the most significant gains from the federation process, consistent with the accuracy results. Centralized learning for RF performs poorly in some cases ($0.499-0.961$), whereas DFL achieves consistently high F1-scores ($0.98$ across all providers), with DFL+Gossip further improving to $0.982$. A similar trend is observed for DT, where centralized learning shows lower performance ($0.672-0.961$), while DFL and CFL provide stable results ($0.967-0.969$), with DFL+Gossip further improving to $0.982$ for AdGuard and Quad9.

\begin{table*}[tbph]
    \centering
    \caption{F1-Score for different DNS providers and validation scenarios. NFL: Non-Federated Learning; CFL: Centralized Federated Learning; DFL: Decentralized Federated Learning; DFL+Gsp:  Gossip-based DFL.}
    \label{tab:performance_metrics_f1}
    
    \resizebox{\textwidth}{!}{%
    \begin{tabular}{|c|c|c|c|c|c|c|c|c|c|c|c|c|c|c|c|c|}
        \hline
        \multirow{2}{*}{\textbf{DNS Provider}} & \multicolumn{4}{c|}{\textbf{SVM}} & \multicolumn{4}{c|}{\textbf{LR}} & \multicolumn{4}{c|}{\textbf{RF}} & \multicolumn{4}{c|}{\textbf{DT}} \\ \cline{2-17}
        & \textbf{NFL} & \textbf{CFL} & \textbf{DFL} & \textbf{DFL+Gsp} & \textbf{NFL} & \textbf{CFL} & \textbf{DFL} & \textbf{DFL+Gsp} & \textbf{NFL} & \textbf{CFL} & \textbf{DFL} & \textbf{DFL+Gsp} & \textbf{NFL} & \textbf{CFL} & \textbf{DFL} & \textbf{DFL+Gsp} \\ \hline
GoogleDNS & 0.940 & 0.969 & 0.969 & 0.960 & 0.849 & 0.93 & 0.93 & 0.849 & 0.961 & 0.985 & 0.98 & 0.982 & 0.958 & 0.961 & 0.967 & 0.979 \\
Cloudflare & 0.945 & 0.969 & 0.969 & 0.965 & 0.893 & 0.93 & 0.93 & 0.945 & 0.612 & 0.985 & 0.98 & 0.982 & 0.672 & 0.961 & 0.967 & 0.981 \\
AdGuard & 0.879 & 0.969 & 0.969 & 0.975 & 0.841 & 0.93 & 0.93 & 0.895 & 0.499 & 0.985 & 0.98 & 0.982 & 0.785 & 0.961 & 0.967 & 0.982 \\
Quad9 & 0.953 & 0.969 & 0.969 & 0.972 & 0.897 & 0.93 & 0.93 & 0.898 & 0.961 & 0.985 & 0.98 & 0.982 & 0.961 & 0.961 & 0.967 & 0.982\\ \hline
    \end{tabular}
    }
\end{table*}

\begin{figure*}[tbph]
    \centering
    \includegraphics[width=\textwidth]{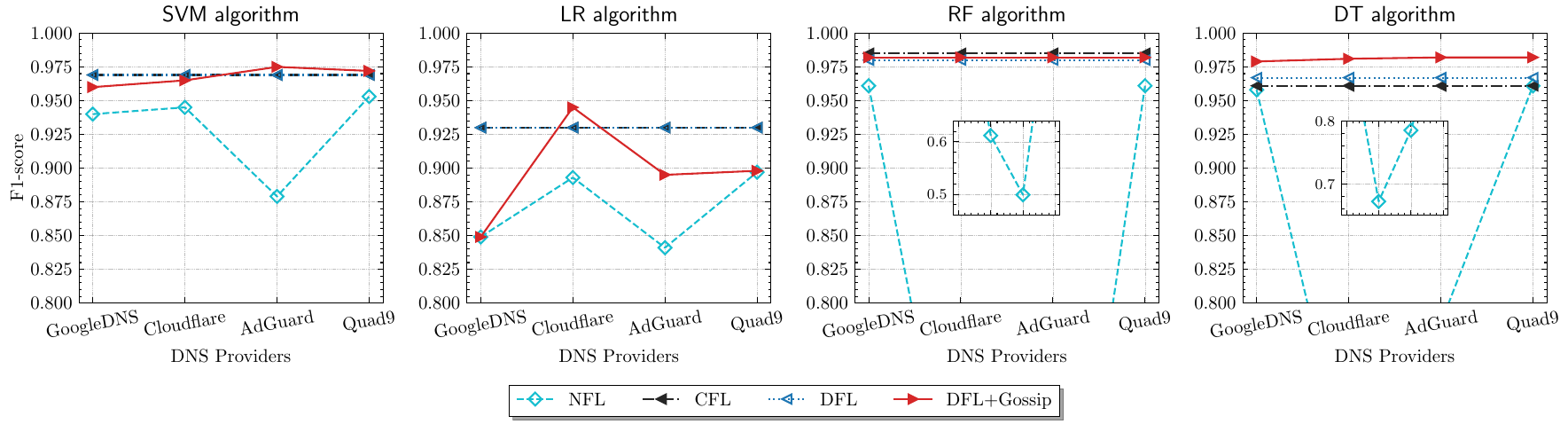}
    \caption{F1-Score for different DNS providers and validation scenarios NFL: Non-Federated Learning; CFL: Centralized Federated Learning; DFL: Decentralized Federated Learning; DFL+Gsp:  Gossip-based DFL.}
    \label{fig:results_line_f1_2}
\end{figure*}

These results, as shown in Figure~\ref{fig:results_line_f1_2}, underscore the significance of decentralized collaboration, especially for models like tree-based algorithms that benefit from heterogeneous decision patterns for improved performance.  The significance of these F1-score improvements is critical in DoH tunnel detection, where precision and recall must be balanced to minimize false positives and false negatives. Moreover, the higher performance of DFL+Gossip compared to standard DFL suggests that incorporating peer-to-peer communication enhances model aggregation and convergence.

In addition to the positive points highlighted in the previous analysis, the main advantage of the DFL+Gossip scenario is the bandwidth savings in communication when sharing models. Assume $n$ represents the number of nodes and $B$ the data volume of model parameters. In CFL, the communication cost is $2nB$, which increases to $n(n-1)B$ in the full DFL scenario. However, by incorporating the Gossip information dissemination mechanism, the cost is reduced to $nB$. Therefore, in this case, even reducing the communication cost by one third with respect to DFL, the results of the proposal maintain or improve the results obtained (except when using LR).

\subsection{Comparative analysis of ML algorithms in related work}
\label{subsec:ml_comparison}

Although earlier sections present individual model metrics, this comparative analysis further highlights the relative merits of each approach within federated scenarios. The RF model consistently outperforms all other evaluated algorithms. Looking at the same results shown in Figure \ref{fig:heatmap_2}, LR algorithm demonstrates the lowest performance among the evaluated models, likely due to its limited ability to capture non-linear relationships inherent in the dataset. It is followed by SVM which, although it obtains acceptable results, does not exceed the results of the others in the DFL+Gossip scenario. Finally, the algorithms based on DT and RF trees stand out, with RF being the only one capable of obtaining very good results in the three federation scenarios. 

In addition to the comparisons made in this work, we also compare the results with other similar works. We will take as a reference the work mentioned in Section~\ref{sec:related_work}~\cite{li2022detecting}, since it is the only work found in the literature that uses FL on the same dataset, although a direct comparison is not possible since it is not the same validation scenario. 

In this paper, a CNN network is used for the training process in CFL, obtaining final accuracy results around $0.9986$ under no-IID data distribution. With our results, we observe that these results are quite similar despite the decentralization of the data and the learning process. In addition, our results have been obtained using computationally simpler algorithms compared to the CNN, demonstrating that the dataset used for validation does not necessitate complex ML models.

The following analysis shows the difference in computational complexity between the ML algorithms used in this comparative study. Let $n$ be the number of training samples, $f$ for the features, and $T$ the number of decision trees, their computational complexities for SVM, LR, DT and RF are $\mathcal{O}(nf)$, $\mathcal{O}(nf)$, $\mathcal{O}(n\log(n)f)$ and $\mathcal{O}(Tn\log(n)f)$, respectively~\cite{Bishop06}. On the other hand, the CNN architecture used in the cited work includes two convolutional layers, two pooling layers, three fully connected layers and one output layer. Given that $k$ is the filter size, $c$ is the number of channels, $L$ is the number of fully connected layers, and $h$ is the number of neurons per layer, the training computational complexity (adapted for feature datasets and no images) can be expressed as $\mathcal{O}(nfkc+Lh^2)$, which is significantly higher than those of the previously mentioned models. While this complexity is significantly higher during training—primarily due to the fully connected layers—CNNs may offer better scalability for larger problems. Once trained, CNNs can also achieve faster inference times compared to models like SVM. Future work should consider evaluating CNN performance under different validation scenarios to more clearly define its advantages in our contributions and objectives.

\subsection{Scalability analysis of CO-DEFEND}
\label{subsec:scalability_analysis}

Beyond the theoretical scalability analysis in Section~\ref{subsec:convergence_analysis}, this subsection experimentally analyzes how our proposal, CO-DEFEND, scales with the number of participating nodes—a key factor in decentralized systems. A primary challenge in designing this experiment was the inherent structure of the CIRA-CIC-DoHBrw-2020 dataset. An initial analysis revealed a considerable class imbalance, with malicious samples significantly outnumbering benign ones by a margin of over 12-to-1. The scarcity of benign data, particularly for certain providers, complicates the creation of a large number of statistically meaningful client partitions for scalability studies. To address this challenge and build a robust foundation for analysis, we first developed a data augmentation pipeline to generate high-fidelity synthetic benign samples.

This pipeline employed a provider-specific approach, where a suite of augmentation techniques (Gaussian Copulas, KDE, SMOTE, and MixUp) were evaluated for each of the four DNS providers. The optimal technique was selected based on a comprehensive set designed to meassure the statistical fidelity. We assessed the preservation of statistical properties using multiple divergence metrics: (i) Wasserstein distance tests to measure the similarity of marginal distributions for each feature, and (ii) Maximum Mean Discrepancy (MMD) and a Classifier Two-Sample Test (C2ST) to evaluate the indistinguishability of the joint distributions of real and synthetic data. The analysis identified MixUp as the most effective method for AdGuard, Cloudflare, and GoogleDNS, while Kernel Density Estimation (KDE) was chosen for Quad9. This process yielded a robust and balanced dataset, enabling a meaningful scalability evaluation.

To assess performance during scalability testing, we considered the accuracy metric, which will be evaluated in different communication topologies in DFL+Gossip scenario. The experiment evaluates the scenario across communication topologies of 4, 8, 12, and 16 nodes. For scenarios with more than four nodes, additional clients were simulated by partitioning each of the four augmented provider datasets into an equal number of IID shards. For instance, the 8-node topology consists of two nodes for each original provider (e.g., two AdGuard nodes, two Cloudflare nodes, etc.), created by randomly splitting the provider's augmented data in half. We also varied the Gossip model-sharing probability between $p=0.1$ and $p=0.5$ (the same proportion considered until now) and have not evaluated higher values, as this approaches a fully-connected topology. These probabilities were selected to represent network connectivities from sparse to moderately dense, capturing how increasing the fraction of shared peers accelerates convergence. Going beyond this range would result in higher communication overhead while offering only marginal performance improvements, which is contrary to the probabilistic, peer-to-peer nature of gossip protocols.

\begin{figure}[tbph]
    \centering
    \includegraphics[width=1.0\textwidth]{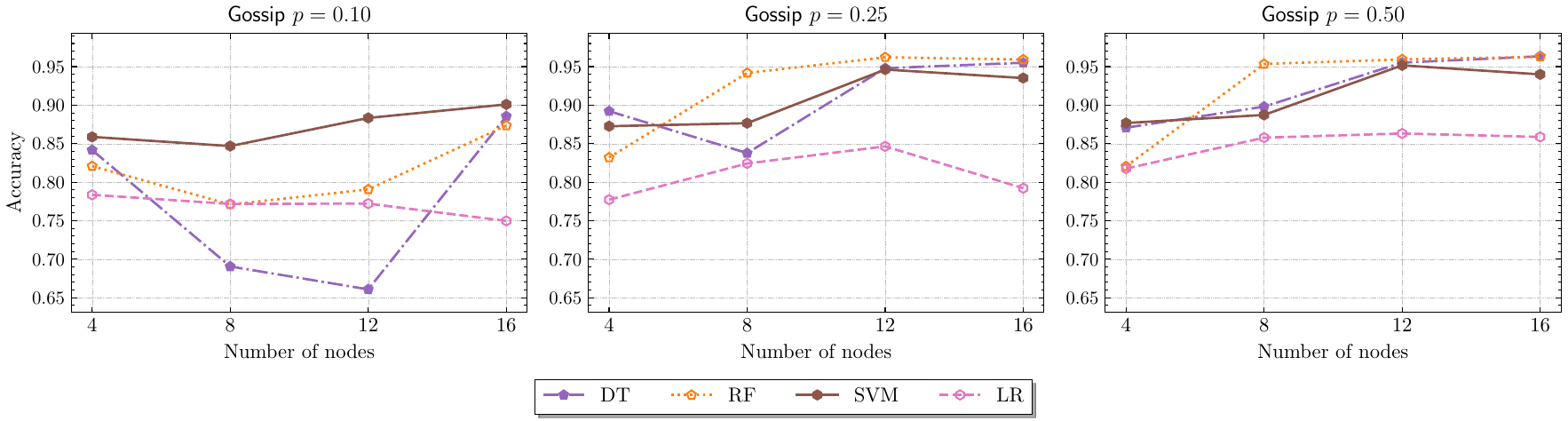}
    \caption{Evaluation of general accuracy performance under different communication topologies (number of nodes and probability of model sharing). Each accuracy value is obtained as an average of all nodes in the topology.}
    \label{fig:scalability_average_nodes}
\end{figure}

As illustrated in Figure~\ref{fig:scalability_average_nodes}, we have evaluated the accuracy performance by considering the average of all nodes in the topology to analyze the general behavior. When considering a Gossip probability of $p=0.1$, no clear tendency is evident across different ML models. For instance, the SVM and RF algorithms demonstrate limited improvements with increased number of nodes, while the LR algorithm experiences slight deterioration. This phenomenon can be explained by the fact that only a limited number of neighbor models are received in each round (between 1-2). As a result, it is logical that learning practically maintains the performance. It is evident from the data that as the Gossip probability is increased from $p=0.25$ to $p=0.5$, there is a notable performance convergence, a trend that is further pronounced when the number of nodes is increased. This can also be explained as the mean number of received models varies between 4-8 when the number of nodes is 16, accelerating the collaborative training.

\begin{figure}[tbph]
    \centering
    \includegraphics[width=1.0\textwidth]{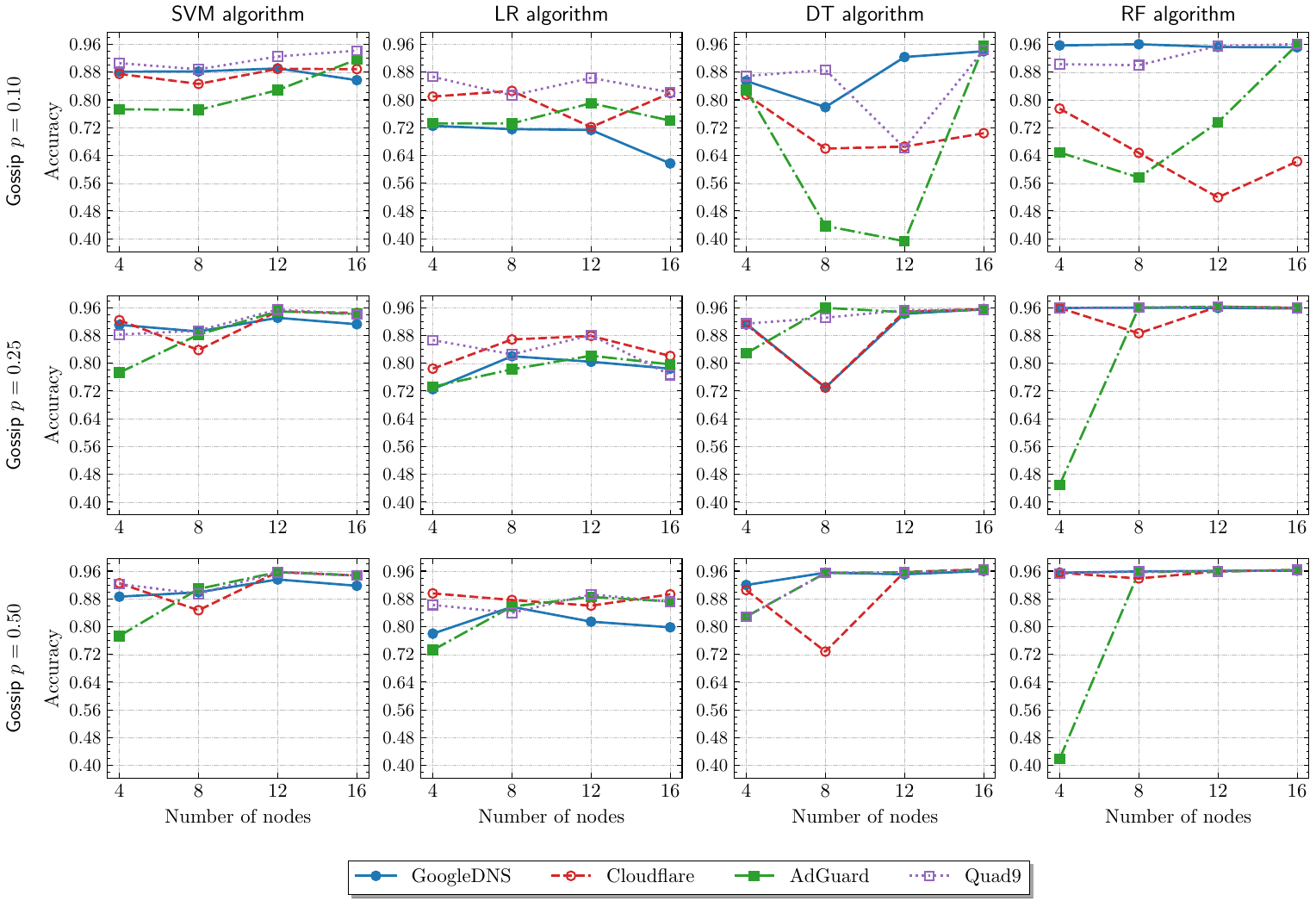}
    \caption{Evaluation of per-provider accuracy performance under different communication topologies (number of nodes and probability of model sharing). Each accuracy value is obtained as an average of the nodes with a sub-dataset from the same provider.}
    \label{fig:scalability_average_per_provider}
\end{figure}

To clearly identify how these parameters affect the framework in a fine-grained way, we have also considered the scalability analysis by analyzing the performance of different DNS providers. In this case, the accuracy performance is now evaluated by grouping and averaging the results from the different providers, i.e., nodes with the same sample origin. As illustrated in Figure~\ref{fig:scalability_average_per_provider}, when $p=0.1$, results are comparable to the previous scenario. The number of shared models is low, learning convergence is not evident, and each group of providers exhibits distinct performance outcomes. In fact, the DT algorithm is the most affected algorithm because it works by extrapolating branches from received models that adapt properly to its own local model (in this case, two models are received at most). As demonstrated in the heterogeneity analysis in Section~\ref{subsec:heterogeinity_analysis}, the quality of the data is less important than the effectiveness of the rules that generalize well to specific cases. For instance, Cloudflare and Ad Guard had fewer samples, so they had to wait until other tree rules fit well in their own distribution, while GoogleDNS and Quad9 obtained good results in general. However, the RF algorithm demonstrates superior performance in comparison to the DT algorithm due to the fact that the RF algorithm utilizes a greater number of trees within the forest, although the convergence rate is more gradual. On the other hand, as the Gossip probability increases to 0.5, most of the providers with the different ML algorithms achieve performance convergence since the number of received models also increased. 

Finally, we have analyzed the communication costs incurred with the different ML models, as it is also an essential study in decentralized environments. For parametric ML models such as SVM and LR, the transmitted parameter vectors are extremely lightweight, requiring only 644 bytes per model share (including other information for coordinating). Due to their structural representation, both the DT and RF models are larger, with an average tree model size of 0.17 MB. In contrast, state-of-the-art deep neural networks impose substantially higher communication costs. For example, the well-known ResNet-18 model requires up to 45 MB for image classification. According to the specifications provided, the described convolutional neural network model in~\cite{li2022detecting} has a model size of approximately 0.887 MB. Furthermore, our designed framework incorporates lightweight Gossip protocol, enabling us to significantly reduce the total amount of shared information to a fraction of $p$, yielding comparable results, as demonstrated in the previous subsection. Consequently, our gossip-based approach achieves a favorable balance between collaborative performance gains and gains in communication efficiency.

\section{CONCLUSION}
\label{sec:conclusion}

The use of DoH by an attacker to conceal malicious activities within encrypted DNS traffic poses a significant challenge to traditional network security approaches. While centralized machine learning models can be effective in detecting DoH covert channels, these models require the collection of DNS traffic from multiple sources, raising privacy and security concerns. Additionally, the centralized approach may result in congestion and single points of failure, limiting its effectiveness in large-scale deployments.

The main contribution of this work lies in the proposal, CO-DEFEND, a Gossip-based DFL collaborative system. DFL offers a privacy-preserving alternative by allowing multiple entities to train a global model without sharing raw data. Unlike traditional FL, which relies on a central node for model aggregation, DFL employs peer-to-peer updates, eliminating the need for centralized coordination. In addition, the batch-based continuous training approach allows adaptation to dynamic environments where a complete dataset is not available from the start.

Another relevant contribution of this work is the adaptation and application of classical machine learning algorithms, such as SVM and RF, to the federation process. Compared to neural networks, which dominate most of the studies in this field, these algorithms (and therefore, the proposed Gossip-based DFL framework) present a lower computational complexity in the training phase, achieving comparable results in many cases.

Our experiments show significant improvements in accuracy and F1-score of trained models compared to distributed scenarios without collaborative learning. Furthermore, the proposal outperforms the results obtained in CFL schemes, in addition to offering advantages inherent to the DFL approach, such as higher scalability, single-node fault tolerance, and reduced communication costs thanks to efficient dissemination algorithms based on Gossip. Despite the use of PCA potentially reducing accuracy, the results have shown improvements over centralized approaches while providing benefits by mitigating overfitting issues and risk-a reduction step that make benign–malicious mimicry more constrained in practice.

The primary threat to the external validity of our findings is the reliance on a single public dataset, CIRA-CIC-DoHBrw-2020. This is a consequence of the current and widely acknowledged scarcity of public data for the specific task of malicious DoH tunneling. While this approach ensures reproducibility and comparability against the established benchmark, the performance of CO-DEFEND may vary on network traffic with different underlying characteristics or against novel tunneling techniques not represented in the CIRA-CIC-DoHBrw-2020 corpus. We thus highlight the community's need for new, diverse, and publicly available DoH datasets to further validate and advance research in this area.

One of the critical challenges in DFL for malicious DoH tunnel classification is the presence of adversarial participants seeking to compromise the integrity of the global model. By removing the central server, malicious nodes could inject manipulated updates to degrade the performance of the model, misclassify malicious DoH traffic as benign, or introduce backdoors. Our present evaluation assumes benign participants and relies only on validation-based selection/pruning for tree-based models, which is not a full Byzantine-defense. To mitigate this, blockchain-based verification can be integrated into the DFL framework to ensure model integrity and model poisoning detection. Additionally, CO-DEFEND is also compatible with secure aggregation and differential privacy in order to assess privacy concerns., and a thorough assessment of their privacy–utility–overhead trade-offs is left to future work. Future work should evaluate the trade-offs between security/privacy, computational overhead, and communication efficiency when implementing these defenses.

On the other hand, although our proposal's batch-based continual training approach for the DFL scheme allows to handle gradual changes in traffic behavior (evolving benign or malicious DNS patterns), its performance under abrupt or recurring drifts, such as the sudden emergence/disappearance of distinct attack families, has not yet been validated. Thus, the validation and design of techniques to alleviate the possible effect of catastrophic forgetting and concept drift phenomenons in challenging incremental learning scenarios are proposed as a line of future work.

\section*{Acknowledgment}

This work was supported by the grant COMPROMISE (PID2020-113795RB-C32 and PID2020-113795RB-C33) funded by MICIU/AEI/10.13039/501100011033; the grant DISCOVERY (PID2023-148716OB-C31 and PID2023-148716OB-C33) funded by MCIU/AEI/10.13039/501100011033 and FEDER, UE; the Grant QURSA (TED2021-130369B and TED2021-130369B-C32) funded by MICIU/AEI/10.13039/501100011033 and European Union Next-GenerationEU/PRTR and it also has been funded by the Galician Regional Government under project ED431B 2024/41 (GPC).
In part, this work was also carried out within the framework of funding provided by the Recovery, Transformation and Resilience Plan, financed by the European Union (Next Generation), partly through the ``TRUFFLES: TRUsted Framework for Federated LEarning Systems'' Project, within the strategic cybersecurity projects (INCIBE, Spain), and partly through the I-Shaper Project (strategic project C114/23).



\begin{thebibliography}{10}
\providecommand{\url}[1]{#1}
\csname url@samestyle\endcsname
\providecommand{\newblock}{\relax}
\providecommand{\bibinfo}[2]{#2}
\providecommand{\BIBentrySTDinterwordspacing}{\spaceskip=0pt\relax}
\providecommand{\BIBentryALTinterwordstretchfactor}{4}
\providecommand{\BIBentryALTinterwordspacing}{\spaceskip=\fontdimen2\font plus
\BIBentryALTinterwordstretchfactor\fontdimen3\font minus \fontdimen4\font\relax}
\providecommand{\BIBforeignlanguage}[2]{{%
\expandafter\ifx\csname l@#1\endcsname\relax
\typeout{** WARNING: IEEEtran.bst: No hyphenation pattern has been}%
\typeout{** loaded for the language `#1'. Using the pattern for}%
\typeout{** the default language instead.}%
\else
\language=\csname l@#1\endcsname
\fi
#2}}
\providecommand{\BIBdecl}{\relax}
\BIBdecl

\bibitem{hoffman2018dns}
P.~E. Hoffman and P.~McManus, ``{DNS Queries over HTTPS (DoH)},'' RFC 8484, 2018.

\bibitem{bumanglag2020impact}
K.~Bumanglag and H.~Kettani, ``{On the impact of DNS over HTTPS paradigm on cyber systems},'' in \emph{3rd International Conference on Information and Computer Technologies (ICICT)}.\hskip 1em plus 0.5em minus 0.4em\relax IEEE, 2020, pp. 494--499.

\bibitem{wang2021comprehensive}
Y.~Wang, A.~Zhou, S.~Liao, R.~Zheng, R.~Hu, and L.~Zhang, ``{A comprehensive survey on DNS tunnel detection},'' \emph{Computer Networks}, vol. 197, 2021.

\bibitem{montazeri2020detection}
M.~MontazeriShatoori, L.~Davidson, G.~Kaur, and A.~Habibi~Lashkari, ``Detection of {DoH} tunnels using time-series classification of encrypted traffic,'' in \emph{IEEE Intl. Conf. on Dependable, Autonomic and Secure Computing (DASC)}, 2020, pp. 63--70.

\bibitem{McMahan17}
H.~B. McMahan, E.~Moore, D.~Ramage, S.~Hampson, and B.~A. y~Arcas, ``Communication-efficient learning of deep networks from decentralized data,'' \emph{Proceedings of the 20 th International Conference on Artificial Intelligence and Statistics (AISTATS) 2017. JMLR: W\&CP volume 54}, 2016, doi: 10.48550/arXiv.1602.05629.

\bibitem{Hallaji24}
E.~Hallaji, R.~Razavi-Far, M.~Saif, B.~Wang, and Q.~Yang, ``Decentralized federated learning: A survey on security and privacy,'' \emph{IEEE Transactions on Big Data}, vol.~10, no.~2, pp. 194--213, 2024.

\bibitem{mockapetris1987domain}
P.~Mockapetris, ``{Domain names - implementation and specification},'' RFC 1035, 1987.

\bibitem{hynek2022summary}
K.~Hynek, D.~Vekshin, J.~Luxemburk, T.~Cejka, and A.~Wasicek, ``Summary of {DNS} over {HTTPS} abuse,'' \emph{IEEE Access}, vol.~10, pp. 54\,668--54\,680, 2022.

\bibitem{merlo2011comparative}
A.~Merlo, G.~Papaleo, S.~Veneziano, and M.~Aiello, ``{A Comparative Performance Evaluation of DNS Tunneling Tools},'' in \emph{Computational Intelligence in Security for Information Systems}.\hskip 1em plus 0.5em minus 0.4em\relax Springer, 2011, pp. 84--91.

\bibitem{iodine}
\BIBentryALTinterwordspacing
E.~Ekman and B.~Andersson, ``iodine,'' 2014. [Online]. Available: \url{https://github.com/yarrick/iodine}
\BIBentrySTDinterwordspacing

\bibitem{dns2tcp}
\BIBentryALTinterwordspacing
O.~Dembour and N.~Collignon, ``dns2tcp,'' 2017. [Online]. Available: \url{https://github.com/alex-sector/dns2tcp}
\BIBentrySTDinterwordspacing

\bibitem{dnscat2}
\BIBentryALTinterwordspacing
R.~Bowes, ``dnscat2,'' 2017. [Online]. Available: \url{https://github.com/iagox86/dnscat2}
\BIBentrySTDinterwordspacing

\bibitem{schmid2021thirty}
G.~Schmid, ``{Thirty years of DNS insecurity: Current issues and perspectives},'' \emph{IEEE Communications Surveys \& Tutorials}, vol.~23, no.~4, pp. 2429--2459, 2021.

\bibitem{turing2019analysis}
A.~Turing and G.~Ye, ``{An Analysis of Godlua Backdoor},'' \emph{360 Netlab Blog}, 2019.

\bibitem{apt}
\BIBentryALTinterwordspacing
C.~Cimpanu, ``{Iranian hacker group becomes first known APT to weaponize DNS-over-HTTPS (DoH)},'' 2020. [Online]. Available: \url{https://zd.net/3EBD8OS}
\BIBentrySTDinterwordspacing

\bibitem{Bittencourt18}
\BIBentryALTinterwordspacing
L.~Bittencourt, R.~Immich, R.~Sakellariou, N.~Fonseca, E.~Madeira, M.~Curado, L.~Villas, L.~DaSilva, C.~Lee, and O.~Rana, ``The internet of things, fog and cloud continuum: Integration and challenges,'' \emph{Internet of Things}, vol. 3-4, pp. 134--155, 2018. [Online]. Available: \url{https://www.sciencedirect.com/science/article/pii/S2542660518300635}
\BIBentrySTDinterwordspacing

\bibitem{Wen23}
J.~Wen, Z.~Zhang, Y.~Lan, Z.~Cui, J.~Cai, and W.~Zhang, ``A survey on federated learning: challenges and applications,'' \emph{International Journal of Machine Learning and Cybernetics}, vol.~14, no.~2, pp. 513--535, 2023.

\bibitem{Mothukuri21}
V.~Mothukuri, R.~M. Parizi, S.~Pouriyeh, Y.~Huang, A.~Dehghantanha, and G.~Srivastava, ``A survey on security and privacy of federated learning,'' \emph{Future Generation Computer Systems}, vol. 115, pp. 619--640, 2021.

\bibitem{Fang20}
M.~Fang, X.~Cao, J.~Jia, and N.~Gong, ``Local model poisoning attacks to $\{$Byzantine-Robust$\}$ federated learning,'' in \emph{29th USENIX security symposium (USENIX Security 20)}, 2020, pp. 1605--1622.

\bibitem{Shi22}
J.~Shi, W.~Wan, S.~Hu, J.~Lu, and L.~Y. Zhang, ``Challenges and approaches for mitigating byzantine attacks in federated learning,'' in \emph{2022 IEEE International Conference on Trust, Security and Privacy in Computing and Communications (TrustCom)}.\hskip 1em plus 0.5em minus 0.4em\relax IEEE, 2022, pp. 139--146.

\bibitem{Schlegel23}
R.~Schlegel, S.~Kumar, E.~Rosnes, and A.~G. i~Amat, ``Codedpaddedfl and codedsecagg: Straggler mitigation and secure aggregation in federated learning,'' \emph{IEEE Transactions on Communications}, vol.~71, no.~4, pp. 2013--2027, 2023.

\bibitem{Geiping20}
J.~Geiping, H.~Bauermeister, H.~Dr{\"o}ge, and M.~Moeller, ``Inverting gradients-how easy is it to break privacy in federated learning?'' \emph{Advances in neural information processing systems}, vol.~33, pp. 16\,937--16\,947, 2020.

\bibitem{Nasr19}
M.~Nasr, R.~Shokri, and A.~Houmansadr, ``Comprehensive privacy analysis of deep learning: Passive and active white-box inference attacks against centralized and federated learning,'' in \emph{2019 IEEE symposium on security and privacy (SP)}.\hskip 1em plus 0.5em minus 0.4em\relax IEEE, 2019, pp. 739--753.

\bibitem{elgabli2025novel}
A.~Elgabli and W.~Mesbah, ``A novel approach for differential privacy-preserving federated learning,'' \emph{IEEE Open Journal of the Communications Society}, vol.~6, pp. 466--476, 2025.

\bibitem{Yuan24}
L.~Yuan, Z.~Wang, L.~Sun, S.~Y. Philip, and C.~G. Brinton, ``Decentralized federated learning: A survey and perspective,'' \emph{IEEE Internet of Things Journal}, 2024.

\bibitem{Ye22}
H.~Ye, L.~Liang, and G.~Y. Li, ``Decentralized federated learning with unreliable communications,'' \emph{IEEE journal of selected topics in signal processing}, vol.~16, no.~3, pp. 487--500, 2022.

\bibitem{Boyd06}
S.~Boyd, A.~Ghosh, B.~Prabhakar, and D.~Shah, ``Randomized gossip algorithms,'' \emph{IEEE transactions on information theory}, vol.~52, no.~6, pp. 2508--2530, 2006.

\bibitem{cajaraville2024byzantine}
D.~Cajaraville-Aboy, A.~Fern{\'a}ndez-Vilas, R.~P. D{\'\i}az-Redondo, and M.~Fern{\'a}ndez-Veiga, ``Byzantine-robust aggregation for securing decentralized federated learning,'' \emph{arXiv preprint arXiv:2409.17754}, 2024.

\bibitem{banadaki2020detecting}
Y.~M. Banadaki, ``Detecting malicious {DNS} over {HTTPS} traffic in domain name system using machine learning classifiers,'' \emph{Journal of Computer Sciences and Applications}, vol.~8, no.~2, 2020.

\bibitem{singh2020detecting}
S.~K. Singh and P.~K. Roy, ``Detecting malicious {DNS} over {HTTPS} traffic using machine learning,'' in \emph{2020 International Conference on Innovation and Intelligence for Informatics, Computing and Technologies (3ICT)}.\hskip 1em plus 0.5em minus 0.4em\relax IEEE, 2020, pp. 1--6.

\bibitem{behnke2021feature}
M.~Behnke, N.~Briner, D.~Cullen, K.~Schwerdtfeger, J.~Warren, R.~Basnet, and T.~Doleck, ``Feature engineering and machine learning model comparison for malicious activity detection in the {DNS}-over-{HTTPS} protocol,'' \emph{IEEE Access}, vol.~9, pp. 129\,902--129\,916, 2021.

\bibitem{alenezi2021classifying}
R.~Alenezi and S.~A. Ludwig, ``Classifying {DNS} tunneling tools for malicious {DoH} traffic,'' in \emph{2021 IEEE Symposium Series on Computational Intelligence (SSCI)}, 2021, pp. 1--9.

\bibitem{jha2021detection}
H.~Jha, I.~Patel, G.~Li, A.~K. Cherukuri, and S.~Thaseen, ``Detection of tunneling in {DNS} over {HTTPS},'' in \emph{2021 7th International Conference on Signal Processing and Communication (ICSC)}, 2021, pp. 42--47.

\bibitem{zebin2022explainable}
T.~Zebin, S.~Rezvy, and Y.~Luo, ``An explainable {AI}-based intrusion detection system for {DNS} over {HTTPS} {(DoH)} attacks,'' \emph{IEEE Transactions on Information Forensics and Security}, vol.~17, pp. 2339--2349, 2022.

\bibitem{abu2023lightweight}
Q.~Abu Al-Haija, M.~Alohaly, and A.~Odeh, ``A lightweight double-stage scheme to identify malicious {DNS} over {HTTPS} traffic using a hybrid learning approach,'' \emph{Sensors}, vol.~23, no.~7, p. 3489, 2023.

\bibitem{alzighaibi2023detection}
A.~R. Alzighaibi, ``Detection of {DoH} traffic tunnels using deep learning for encrypted traffic classification,'' \emph{Computers}, vol.~12, no.~3, p.~47, 2023.

\bibitem{mitsuhashi2023malicious}
R.~Mitsuhashi, Y.~Jin, K.~Iida, T.~Shinagawa, and Y.~Takai, ``Malicious {DNS} tunnel tool recognition using persistent {DoH} traffic analysis,'' \emph{IEEE Transactions on Network and Service Management}, vol.~20, no.~2, pp. 2086--2095, 2023.

\bibitem{niktabe2024detection}
S.~Niktabe, A.~H. Lashkari, and D.~P. Sharma, ``Detection, characterization, and profiling {DoH} malicious traffic using statistical pattern recognition,'' \emph{International Journal of Information Security}, vol.~23, no.~2, pp. 1293--1316, 2024.

\bibitem{niktabe2024unveiling}
S.~Niktabe, A.~H. Lashkari, and A.~H. Roudsari, ``Unveiling {DoH} tunnel: Toward generating a balanced {DoH} encrypted traffic dataset and profiling malicious behavior using inherently interpretable machine learning,'' \emph{Peer-to-Peer Networking and Applications}, vol.~17, no.~1, pp. 507--531, 2024.

\bibitem{domingos2000mining}
P.~Domingos and G.~Hulten, ``Mining high-speed data streams,'' in \emph{Proceedings of the sixth ACM SIGKDD international conference on Knowledge discovery and data mining}, 2000, pp. 71--80.

\bibitem{bifet2010moa}
A.~Bifet, G.~Holmes, B.~Pfahringer, P.~Kranen, H.~Kremer, T.~Jansen, and T.~Seidl, ``Moa: Massive online analysis, a framework for stream classification and clustering,'' in \emph{Proceedings of the first workshop on applications of pattern analysis}.\hskip 1em plus 0.5em minus 0.4em\relax PMLR, 2010, pp. 44--50.

\bibitem{gama2014survey}
J.~Gama, I.~{\v{Z}}liobait{\.e}, A.~Bifet, M.~Pechenizkiy, and A.~Bouchachia, ``A survey on concept drift adaptation,'' \emph{ACM computing surveys (CSUR)}, vol.~46, no.~4, pp. 1--37, 2014.

\bibitem{bifet2007learning}
A.~Bifet and R.~Gavalda, ``Learning from time-changing data with adaptive windowing,'' in \emph{Proceedings of the 2007 SIAM international conference on data mining}.\hskip 1em plus 0.5em minus 0.4em\relax SIAM, 2007, pp. 443--448.

\bibitem{manapragada2018extremely}
C.~Manapragada, G.~I. Webb, and M.~Salehi, ``Extremely fast decision tree,'' in \emph{Proceedings of the 24th ACM SIGKDD International Conference on Knowledge Discovery \& Data Mining}, 2018, pp. 1953--1962.

\bibitem{marfoq2023federated}
O.~Marfoq, G.~Neglia, L.~Kameni, and R.~Vidal, ``Federated learning for data streams,'' in \emph{International Conference on Artificial Intelligence and Statistics}.\hskip 1em plus 0.5em minus 0.4em\relax PMLR, 2023, pp. 8889--8924.

\bibitem{jothimurugesan2023federated}
E.~Jothimurugesan, K.~Hsieh, J.~Wang, G.~Joshi, and P.~B. Gibbons, ``Federated learning under distributed concept drift,'' in \emph{International Conference on Artificial Intelligence and Statistics}.\hskip 1em plus 0.5em minus 0.4em\relax PMLR, 2023, pp. 5834--5853.

\bibitem{wu2024effective}
Y.~Wu, H.~Yang, X.~Wang, H.~Yu, A.~El~Saddik, and M.~S. Hossain, ``An effective federated learning system for industrial iot data streaming,'' \emph{Alexandria Engineering Journal}, vol. 105, pp. 414--422, 2024.

\bibitem{GONZALEZSOTO2024110137}
\BIBentryALTinterwordspacing
M.~González-Soto, R.~P. Díaz-Redondo, M.~Fernández-Veiga, B.~Fernández-Castro, and A.~Fernández-Vilas, ``Decentralized and collaborative machine learning framework for iot,'' \emph{Computer Networks}, vol. 239, p. 110137, 2024. [Online]. Available: \url{https://www.sciencedirect.com/science/article/pii/S1389128623005820}
\BIBentrySTDinterwordspacing

\bibitem{Qayyum22}
A.~Qayyum, K.~Ahmad, M.~A. Ahsan, A.~Al-Fuqaha, and J.~Qadir, ``Collaborative federated learning for healthcare: Multi-modal covid-19 diagnosis at the edge,'' \emph{IEEE Open Journal of the Computer Society}, vol.~3, pp. 172--184, 2022.

\bibitem{Long20}
G.~Long, Y.~Tan, J.~Jiang, and C.~Zhang, ``Federated learning for open banking,'' in \emph{Federated learning: privacy and incentive}.\hskip 1em plus 0.5em minus 0.4em\relax Springer, 2020, pp. 240--254.

\bibitem{Su21}
Z.~Su, Y.~Wang, T.~H. Luan, N.~Zhang, F.~Li, T.~Chen, and H.~Cao, ``Secure and efficient federated learning for smart grid with edge-cloud collaboration,'' \emph{IEEE Transactions on Industrial Informatics}, vol.~18, no.~2, pp. 1333--1344, 2021.

\bibitem{Zhang21}
W.~Zhang, D.~Yang, W.~Wu, H.~Peng, N.~Zhang, H.~Zhang, and X.~Shen, ``Optimizing federated learning in distributed industrial iot: A multi-agent approach,'' \emph{IEEE Journal on Selected Areas in Communications}, vol.~39, no.~12, pp. 3688--3703, 2021.

\bibitem{Doriguzzi24}
R.~Doriguzzi-Corin and D.~Siracusa, ``Flad: adaptive federated learning for ddos attack detection,'' \emph{Computers \& Security}, vol. 137, p. 103597, 2024.

\bibitem{Zhao19}
Y.~Zhao, J.~Chen, D.~Wu, J.~Teng, and S.~Yu, ``Multi-task network anomaly detection using federated learning,'' in \emph{Proceedings of the 10th international symposium on information and communication technology}, 2019, pp. 273--279.

\bibitem{li2022detecting}
B.~Li, S.~He, H.~Peng, E.~Zhang, and J.~Xin, ``Detecting {DoH} tunnels with privacy protection using federated learning,'' in \emph{International Conference on Network Communication and Information Security (ICNCIS 2021)}, vol. 12175.\hskip 1em plus 0.5em minus 0.4em\relax SPIE, 2022, pp. 133--141.

\bibitem{Beltran23}
\BIBentryALTinterwordspacing
E.~T. Martínez~Beltrán, M.~Q. Pérez, P.~M.~S. Sánchez, S.~L. Bernal, G.~Bovet, M.~G. Pérez, G.~M. Pérez, and A.~H. Celdrán, ``Decentralized federated learning: Fundamentals, state of the art, frameworks, trends, and challenges,'' \emph{IEEE Communications Surveys and Tutorials}, vol.~25, no.~4, p. 2983–3013, 2023, doi: 10.1109/comst.2023.3315746. [Online]. Available: \url{http://dx.doi.org/10.1109/COMST.2023.3315746}
\BIBentrySTDinterwordspacing

\bibitem{Pokhrel20}
S.~R. Pokhrel and J.~Choi, ``Federated learning with blockchain for autonomous vehicles: Analysis and design challenges,'' \emph{IEEE Transactions on Communications}, vol.~68, no.~8, pp. 4734--4746, 2020.

\bibitem{Truex19}
S.~Truex, N.~Baracaldo, A.~Anwar, T.~Steinke, H.~Ludwig, R.~Zhang, and Y.~Zhou, ``A hybrid approach to privacy-preserving federated learning,'' in \emph{Proceedings of the 12th ACM workshop on artificial intelligence and security}, 2019, pp. 1--11.

\bibitem{Markovic22}
T.~Markovic, M.~Leon, D.~Buffoni, and S.~Punnekkat, \emph{Random Forest Based on Federated Learning for Intrusion Detection}.\hskip 1em plus 0.5em minus 0.4em\relax IFIP Advances in Information and Communication Technology, 06 2022, pp. 132--144.

\bibitem{Souza20}
L.~A.~C. de~Souza, G.~Antonio F.~Rebello, G.~F. Camilo, L.~C.~B. Guimarães, and O.~C. M.~B. Duarte, ``Dfedforest: Decentralized federated forest,'' in \emph{2020 IEEE International Conference on Blockchain (Blockchain)}, 2020, pp. 90--97.

\bibitem{Wu20}
Y.~Wu, S.~Cai, X.~Xiao, G.~Chen, and B.~Ooi, ``Privacy preserving vertical federated learning for tree-based models,'' \emph{Proceedings of the VLDB Endowment}, vol.~13, pp. 2090--2103, 08 2020.

\bibitem{Yin18}
D.~Yin, Y.~Chen, K.~Ramchandran, and P.~Bartlett, ``Byzantine-robust distributed learning: Towards optimal statistical rates,'' 2018, doi: 10.48550/arXiv.1803.01498.

\bibitem{Blanchard17}
P.~Blanchard, E.~M.~E. Mhamdi, R.~Guerraoui, and J.~Stainer, ``Byzantine-tolerant machine learning,'' 2017, doi: 10.48550/arXiv.1703.02757.

\bibitem{Corea24}
P.~M. Corea, Y.~Liu, J.~Wang, S.~Niu, and H.~Song, ``Explainable ai for comparative analysis of intrusion detection models,'' in \emph{2024 IEEE International Mediterranean Conference on Communications and Networking (MeditCom)}.\hskip 1em plus 0.5em minus 0.4em\relax IEEE, 2024, pp. 585--590.

\bibitem{Cortes95}
\BIBentryALTinterwordspacing
C.~Cortes and V.~Vapnik, ``Support-vector networks,'' \emph{Machine Learning}, vol.~20, no.~3, pp. 273--297, Sep 1995. [Online]. Available: \url{https://doi.org/10.1007/BF00994018}
\BIBentrySTDinterwordspacing

\bibitem{Wright95}
R.~E. Wright, ``Logistic regression,'' in \emph{Reading and understanding multivariate statistics}, L.~G. Grimm and P.~R. Yarnold, Eds.\hskip 1em plus 0.5em minus 0.4em\relax American Psychological Association, 1995, pp. 217--244.

\bibitem{Breiman84}
L.~Breiman, J.~Friedman, R.~A. Olshen, and C.~J. Stone, \emph{Classification and Regression Trees}.\hskip 1em plus 0.5em minus 0.4em\relax Chapman \& Hall, 1984.

\bibitem{Quinlan86}
\BIBentryALTinterwordspacing
J.~R. Quinlan, ``Induction of decision trees,'' \emph{Mach. Learn.}, vol.~1, no.~1, p. 81–106, Mar. 1986. [Online]. Available: \url{https://doi.org/10.1023/A:1022643204877}
\BIBentrySTDinterwordspacing

\bibitem{Domingos00}
P.~Domingos and G.~Hulten, ``Mining high-speed data streams,'' \emph{Proceeding of the Sixth ACM SIGKDD International Conference on Knowledge Discovery and Data Mining}, 11 2002.

\bibitem{Breiman01}
L.~Breiman, ``Random forests,'' \emph{Machine Learning}, vol.~45, pp. 5--32, 10 2001.

\bibitem{yang2022decentralized}
S.~Yang, X.~Zhang, and M.~Wang, ``Decentralized gossip-based stochastic bilevel optimization over communication networks,'' \emph{Advances in neural information processing systems}, vol.~35, pp. 238--252, 2022.

\bibitem{yuan2016convergence}
K.~Yuan, Q.~Ling, and W.~Yin, ``On the convergence of decentralized gradient descent,'' \emph{SIAM Journal on Optimization}, vol.~26, no.~3, pp. 1835--1854, 2016.

\bibitem{xin2021improved}
R.~Xin, U.~A. Khan, and S.~Kar, ``An improved convergence analysis for decentralized online stochastic non-convex optimization,'' \emph{IEEE Transactions on Signal Processing}, vol.~69, pp. 1842--1858, 2021.

\bibitem{liu2022decentralized}
W.~Liu, L.~Chen, and W.~Zhang, ``Decentralized federated learning: Balancing communication and computing costs,'' \emph{IEEE Transactions on Signal and Information Processing over Networks}, vol.~8, pp. 131--143, 2022.

\bibitem{hoffman2021spectral}
C.~Hoffman, M.~Kahle, and E.~Paquette, ``Spectral gaps of random graphs and applications,'' \emph{International Mathematics Research Notices}, vol. 2021, no.~11, pp. 8353--8404, 2021.

\bibitem{feige2005spectral}
U.~Feige and E.~Ofek, ``Spectral techniques applied to sparse random graphs,'' \emph{Random Structures \& Algorithms}, vol.~27, no.~2, pp. 251--275, 2005.

\bibitem{moure2023real}
\BIBentryALTinterwordspacing
M.~Moure-Garrido, C.~Campo, and C.~Garcia-Rubio, ``Real time detection of malicious doh traffic using statistical analysis,'' \emph{Computer Networks}, vol. 234, p. 109910, 2023. [Online]. Available: \url{https://www.sciencedirect.com/science/article/pii/S1389128623003559}
\BIBentrySTDinterwordspacing

\bibitem{bhagoji2018enhancing}
A.~N. Bhagoji, D.~Cullina, C.~Sitawarin, and P.~Mittal, ``Enhancing robustness of machine learning systems via data transformations,'' in \emph{2018 52nd Annual conference on information sciences and systems (CISS)}.\hskip 1em plus 0.5em minus 0.4em\relax IEEE, 2018, pp. 1--5.

\bibitem{alemany2020dilemma}
S.~Alemany and N.~Pissinou, ``The dilemma between data transformations and adversarial robustness for time series application systems,'' \emph{arXiv preprint arXiv:2006.10885}, 2020.

\bibitem{hossin2015review}
M.~Hossin and M.~N. Sulaiman, ``{A review on evaluation metrics for data classification evaluations},'' \emph{{International journal of data mining \& knowledge management process}}, vol.~5, no.~2, 2015.

\bibitem{Riedel24}
\BIBentryALTinterwordspacing
P.~Riedel, L.~Schick, R.~von Schwerin, M.~Reichert, D.~Schaudt, and A.~Hafner, ``Comparative analysis of open-source federated learning frameworks - a literature-based survey and review,'' \emph{International Journal of Machine Learning and Cybernetics}, vol.~15, no.~11, pp. 5257--5278, Nov 2024. [Online]. Available: \url{https://doi.org/10.1007/s13042-024-02234-z}
\BIBentrySTDinterwordspacing

\bibitem{Bishop06}
C.~M. Bishop and N.~M. Nasrabadi, \emph{Pattern recognition and machine learning}.\hskip 1em plus 0.5em minus 0.4em\relax Springer, 2006, vol.~4, no.~4.

\end{thebibliography}
\end{document}